\documentclass[sigconf]{acmart}

\copyrightyear{2026}
\acmYear{2026}
\setcopyright{cc}
\setcctype{by}
\acmConference[KDD 2026] {Proceedings of the 32nd ACM SIGKDD Conference on Knowledge Discovery and Data Mining V.2}{August 9--13, 2026}{Jeju Island, Republic of Korea.}
\acmBooktitle{Proceedings of the 32nd ACM SIGKDD Conference on Knowledge Discovery and Data Mining V.2 (KDD 2026), August 9--13, 2026, Jeju Island, Republic of Korea}
\acmISBN{979-8-4007-2259-2/2026/08}
\acmDOI{10.1145/3770855.3818178}

\AtBeginDocument{%
  }

\usepackage{amsmath}
\usepackage{xspace}
\usepackage{multirow}
\usepackage{pifont}

\newcommand{\NAME}{TimeGS\xspace}
\newcommand{\EncodeBlockName}{2D Variation Feature Extraction\xspace}
\newcommand{\DecodeBlockName}{Multi-Basis Gaussian Kernel Generation\xspace}
\newcommand{\RenderBlockName}{Multi-Period Chronologically Continuous Rasterization\xspace}
\newcommand{\AggregateBlockName}{Channel-Adaptive Aggregation\xspace}
\newcommand{\EncodeBlockNameShort}{2D-VFE\xspace}
\newcommand{\DecodeBlockNameShort}{MB-GKG\xspace}
\newcommand{\RenderBlockNameShort}{MP-CCR\xspace}
\newcommand{\AggregateBlockNameShort}{CAA\xspace}

\begin{document}

\title{Forecasting as Rendering: A 2D Gaussian Splatting Framework for Time Series Forecasting}

\author{Yixin Wang}
\authornote{Equal contribution.}
\affiliation{
  \institution{Tsinghua Shenzhen International
Graduate School, Tsinghua University}
  \city{Shenzhen}
  \country{China}
}
\email{wangyixin25@mails.tsinghua.edu.cn}

\author{Yifan Hu}
\authornotemark[1]
\affiliation{
  \institution{Tsinghua Shenzhen International
Graduate School, Tsinghua University}
  \city{Shenzhen}
  \country{China}
}
\email{huyf25@mails.tsinghua.edu.cn}

\author{Peiyuan Liu}
\authornotemark[1]
\affiliation{
  \institution{Tsinghua Shenzhen International
Graduate School, Tsinghua University}
  \city{Shenzhen}
  \country{China}
}
\email{lpy23@mails.tsinghua.edu.cn}

\author{Naiqi Li}
\authornote{Corresponding author: Naiqi Li and Tao Dai}
\affiliation{
  \institution{Tsinghua Shenzhen International
Graduate School, Tsinghua University}
  \city{Shenzhen}
  \country{China}
}
\email{linaiqi.thu@gmail.com}

\author{Tao Dai}
\authornotemark[2]
\affiliation{
  \institution{College of Computer Science and Software Engineering, Shenzhen University}
  \city{Shenzhen}
  \country{China}
}
\email{daitao.edu@gmail.com}

\author{Shu-Tao Xia}
\affiliation{
  \institution{Tsinghua Shenzhen International
Graduate School, Tsinghua University}
  \city{Shenzhen}
  \country{China}
}
\email{xiast@sz.tsinghua.edu.cn}

\makeatletter
\renewcommand{\authornote}[1]{%
  \if@ACM@anonymous\else
    \g@addto@macro\@authornotes{%
      \stepcounter{footnote}%
      \footnotetext{#1}}%
  \fi}
\makeatother





\begin{abstract}

Time series forecasting remains a challenging problem due to the intricate entanglement of intra-period fluctuations and inter-period trends. 
While recent advances have attempted to reshape 1D sequences into 2D period-phase representations, they suffer from two principal limitations.
Firstly, treating reshaped tensors as static images results in a topological mismatch, as standard spatial operators sever chronological continuity at grid boundaries. Secondly, relying on uniform fixed-size representations allocates modeling capacity inefficiently and fails to provide the adaptive resolution required for compressible, non-stationary temporal patterns.
To address these limitations, we introduce \NAME, a novel framework that fundamentally shifts the forecasting paradigm from regression to 2D generative rendering. 
By reconceptualizing the future sequence as a latent 2D temporal surface, TimeGS utilizes the inherent anisotropy of Gaussian kernels to adaptively model complex variations with flexible geometric alignment.
To realize this, we introduce a \DecodeBlockName (\DecodeBlockNameShort) block that synthesizes kernels from a fixed dictionary to stabilize optimization, and a \RenderBlockName (\RenderBlockNameShort) block that enforces strict temporal continuity across periodic boundaries.
Comprehensive experiments on standard benchmark datasets demonstrate that \NAME attains state-of-the-art or competitive performance.
The code is at \url{https://github.com/yixinwang1/TimeGS}.
\end{abstract}





\maketitle
\newcommand\kddavailabilityurl{https://doi.org/10.5281/zenodo.8475}
\ifdefempty{\kddavailabilityurl}{}{
\begingroup\small\noindent\raggedright\textbf{Resource Availability:}\\
The code is available at \url{\kddavailabilityurl}.
\endgroup
}

\section{Introduction}

\begin{figure}[!t]
    \centering
    \includegraphics[width=\linewidth]{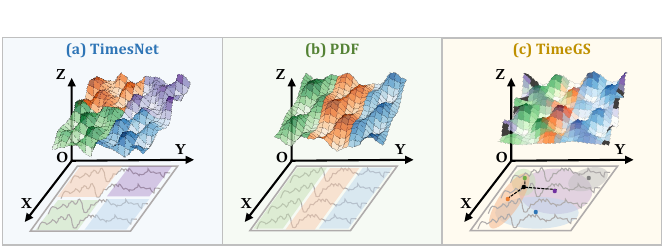}
    \caption{Comparison of 2D Period–based modeling.
    (a) TimesNet \cite{wu2022timesnet} and (b) PDF \cite{pdf} may can break temporal adjacency at row boundaries by using grid-based 2D operators. By contrast, (c) our TimeGS can avoid boundary artifacts by leveraging anisotropic Gaussians and continuous rendering to adapt to information density. (X: period index, Y: phase within a period, Z: value). }
    \vspace{-10pt}
    \label{fig:intro}
\end{figure}


Time series forecasting (TSF) constitutes a fundamental capability for data-driven decision-making in critical domains, including energy management~\cite{deb2017energyreview,salman2024solar}, numerical weather prediction~\cite{karevan2020weather, hewage2020weather} and financial investment~\cite{finmamba, sezer2020financialreview}. 
A key methodological challenge in time series forecasting arises from the need to capture the intricate structural evolutions that inherently exist within continuous real-world data, including both intra-period fluctuations and inter-period trends~\cite{oreshkin2019nbeats, wu2021autoformer, challu2022nhits, wu2022timesnet}.
Recognizing that these evolutions often manifest as recurring cycles,
existing methods exploit temporal periodicity by restructuring a 1D time series into a 2D period-phase representation~\cite{wu2022timesnet, pdf, nematirad2025times2d}. In this way, each row encodes high-frequency fluctuations within a single period, whereas each column aggregates observations at a fixed phase across multiple periods, thereby rendering long-range temporal patterns more explicit. This compact 2D modeling inspires the recent methods of TimesNet \cite{wu2022timesnet} and PDF \cite{pdf} with 2D operators to simultaneously capture local intra-period dynamics and global inter-period evolution, as illustrated in Figure \ref{fig:intro}.


Despite the substantial progress achieved by 2D period-based modeling, these approaches still exhibit two principal limitations. \ding{182} They overlook the fundamental differences in topology and information organization between a reordered time series and a truly 2D spatial lattice. This mismatch can induce systematic biases and characteristic forecasting artifacts. For instance, methods such as TimesNet and PDF employing 2D convolutions may disrupt temporal adjacency at grid boundaries and thereby introduce artificial discontinuities, as illustrated in Figure \ref{fig:intro}, because the rightmost element of one row and the leftmost element of the subsequent row in the 2D layout are temporally consecutive (corresponding to $t$ and $t+1$).
\ding{183} Representing the 2D map using fixed-size windows is inherently restrictive, as real-world time series are frequently compressible: extended stationary segments can be accurately reconstructed from a small number of latent factors, whereas a relatively small subset of change points, anomalous spikes, and phase drifts predominantly determines the forecasting difficulty. Uniform windowing thus allocates modeling capacity inefficiently to redundant regions and fails to provide adaptive resolution and appropriate geometric alignment in the segments that are most critical for prediction.

To address the above issues,  we propose \textbf{\NAME}, a unified framework that adapts the  2D Gaussian Splatting (2DGS) \cite{huang20242dgaussiansplatting} rendering pipeline for time series forecasting. 
Specifically, we reshape 1D historical sequences into 2D tensors and employ multiple \EncodeBlockName (\EncodeBlockNameShort) blocks, which are based on UNet~\cite{ronneberger2015unet}, to extract coupled variation features from diverse temporal views. 
These features are uniformly decoded by the \DecodeBlockName (\DecodeBlockNameShort) block to obtain the shape and intensity of Gaussian kernels, which are subsequently rasterized by the \RenderBlockName (\RenderBlockNameShort) block. Finally, the \AggregateBlockName (\AggregateBlockNameShort) block fuses the forecasts via channel-adaptive weighting.
Furthermore, to address the training instability often associated with optimizing free-floating Gaussian kernels on noisy data, our \DecodeBlockName reformulates shape regression as a stable dictionary learning task using a fixed basis bank.
Moreover, our \RenderBlockName treats Gaussian kernels as continuous signal segments that naturally wrap around the 2D grid boundaries. This ensures that the rendered output maintains strict temporal continuity, reconciling the representational benefits of 2D structural modeling with the sequential nature of time series effectively.

In a nutshell, our contributions are summarized as follows:
\begin{itemize}
\vspace{-3pt}
\item We propose \NAME, pioneering a paradigm shift in time series forecasting from 1D point-wise regression to 2D generative rendering, which explicitly models coupled variations via anisotropic 2D Gaussian splatting.
\item To effectively render time series data, we introduce two critical modules: \DecodeBlockName stabilizes geometric optimization via dictionary learning, and \RenderBlockName resolves the topological mismatch of 2D reshaping.
\item Extensive experiments on widely-used benchmarks demonstrate the state-of-the-art or competitive performance of \NAME, validating that our structural rendering paradigm.
\end{itemize}
\vspace{-3pt}

\begin{figure*}[t!]
  \centering
  \includegraphics[width=\linewidth]{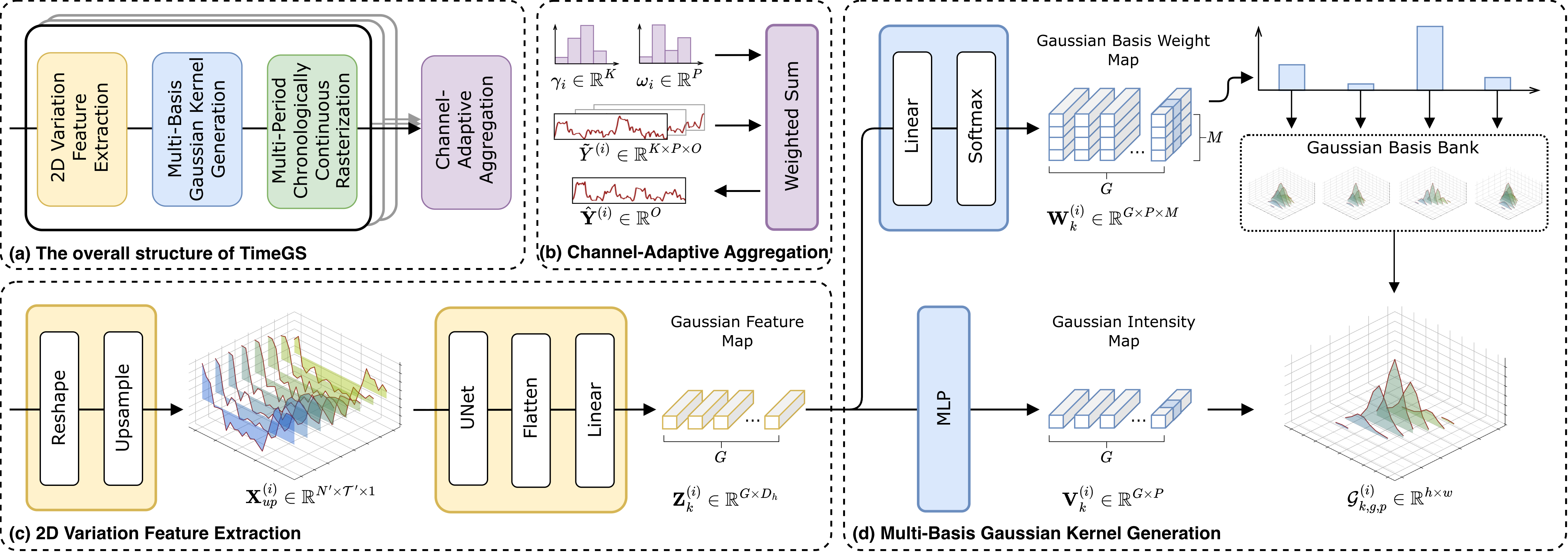}
  \caption{The architecture of \NAME. 
  (a) Overall Structure: The model processes time series through a generative rendering paradigm.
  (b) \AggregateBlockName: Fuses the rendered outputs from different branches using a channel-wise adaptive weighting mechanism to ensure robust multivariate forecasting.
  (c) \EncodeBlockName: Extracts 2D variation features using UNet-based encoder on reshaped temporal tensors.
  (d) \DecodeBlockName: Generates composite Gaussian kernels by predicting intensities and weights for the Gaussian Basis Bank.}
  \vspace{-5pt}
    \label{fig:overview}
\end{figure*}

\section{Related Works}
\subsection{Temporal Variation Modeling}
Deep learning approaches have been widely explored for modeling temporal variations~\cite{duet,timefilter,hu2025bridging}. Transformer-based models~\cite{liu2021pyraformer,zhou2022fedformer, liuitransformer, liu2024timebridge}, such as Informer~\cite{haoyietal-informer-2021} and Autoformer~\cite{wu2021autoformer}, use sparse attention and decomposition for long-range dependencies, while PatchTST~\cite{patchtst} leverages patching and channel independence. MLP-based models~\cite{xu2023fits,cmos,amd,timebase}, such as DLinear~\cite{dlinear} and TimeMixer~\cite{wang2024timemixer}, show that simple temporal structures can remain competitive.

While the above methods excel at modeling global dependencies or performing trend-seasonality decomposition, the 1D perspective often limits their ability to fully resolve the structural nuances of how periodic patterns evolve locally and globally. To address this, recent works explore alternative representations: MICN~\cite{wang2023micn} employs multi-scale isometric convolutions for hierarchical feature extraction, while TimesNet~\cite{wu2022timesnet} reshapes 1D sequences into 2D tensors to explicitly model intra-period and inter-period variations. Different from these methods that mainly apply standard 2D operators on reshaped tensors, our method treats the 2D representation as a periodic temporal surface and performs chronologically continuous Gaussian rendering.

\subsection{Gaussian Splatting}
3D Gaussian Splatting (3DGS)~\cite{kerbl20233dgaussiansplattin} enables high-fidelity rendering by explicitly representing scenes as anisotropic 3D ellipsoids. To improve geometric accuracy and surface alignment, 2D Gaussian Splatting (2DGS) ~\cite{huang20242dgaussiansplatting} was proposed, which flattens the 3D ellipsoids into oriented 2D disks. 

Recent studies leverage 2DGS for image processing tasks.
GaussianImage ~\cite{zhang2024gaussianimage} and GaussianToken ~\cite{dong2025gaussiantoken} show that 2D Gaussian kernels can serve as compact image codes and as discrete visual tokens for downstream tasks.
Furthermore, 2DGS shows strong potential in reconstruction tasks akin to forecasting: GaussianSR ~\cite{hu2025gaussiansr} utilizes 2DGS for arbitrary-scale super-resolution, showing its capability to recover high-frequency details from low-resolution inputs. ~\citet{li2025gsinpaint} explored 2DGS for image inpainting with semantic alignment, demonstrating that Gaussian kernels can effectively fill missing regions based on learned context.

\section{Preliminaries}
\subsection{Problem Statement}
Given a historical time series $\mathbf{X} = [x_1, \dots, x_I] \in \mathbb{R}^{I \times C}$, where $I$ is the look-back window size and $C$ is the number of channels, the goal of TSF is to predict the future sequence $\mathbf{Y} = [y_{1}, \dots, y_{O}] \in \mathbb{R}^{O \times C}$, where $O$ is the forecasting horizon.

\subsection{2D Gaussian Splatting}
2D Gaussian Splatting represents a scene or signal distribution using a set of 2D Gaussian kernels. Each kernel $\mathcal{N}_j$ is characterized by its center position $\mu_j \in \mathbb{R}^2$ and a covariance matrix $\Sigma_j \in \mathbb{R}^{2 \times 2}$. The contribution of the $j$-th branch at any 2D coordinate $\mathbf{p}$ is defined by the probability density function (ignoring the normalization constant for rendering purposes):

\begin{align}
\mathcal{N}_j(\mathbf{p}) = \exp \left( -\frac{1}{2} (\mathbf{p} - \mu_j)^T \Sigma_j^{-1} (\mathbf{p} - \mu_j) \right).
\end{align}

To ensure that the covariance matrix $\Sigma_j$ remains valid (i.e., positive semi-definite) throughout the optimization process, it is not optimized directly. Instead, it is parameterized via Cholesky decomposition. Specifically, we define a lower triangular matrix $\mathbf{L}_j \in \mathbb{R}^{2 \times 2}$:

\begin{align}
\mathbf{L}_j = \begin{bmatrix} l_{11} & 0 \\ l_{21} & l_{22} \end{bmatrix}.
\end{align}

The covariance matrix is then constructed as:
\begin{align}
\Sigma_j = \mathbf{L}_j \mathbf{L}_j^T.
\end{align}

This formulation ensures that $\Sigma_j$ is positive definite, allowing the Gaussian kernel to represent anisotropic and oriented shapes.
In standard splatting rendering, the final pixel value is computed as a weighted sum of all the overlapping Gaussian kernels that cover it.

\section{Methodology}

\subsection{Structure Overview}
The overall architecture of \NAME, as illustrated in Figure \ref{fig:overview}, consists of four sequential components: 
(1) \textbf{\EncodeBlockName (\EncodeBlockNameShort)}, which reshapes time series into 2D tensors and extracts coupled intra-period and inter-period variation features using multiple encoders; 
(2) \textbf{\DecodeBlockName (\DecodeBlockNameShort)}, which dynamically composes kernels from a fixed basis dictionary and predicts the intensity of Gaussian kernels; 
(3) \textbf{\RenderBlockName (\RenderBlockNameShort)}, which resolves the chronological discontinuity problem by converting 2D kernel support into a temporally continuous 1D rendering process across period boundaries;
(4) \textbf{\AggregateBlockName (\AggregateBlockNameShort)}, which fuses the multi-branch forecasts using learnable channel-specific weights to explicitly capture variable-dependent temporal dynamics.

To accommodate the diverse temporal characteristics inherent in multivariate time series, where variables may exhibit distinct dominant periods, or manifest disparate variation patterns even under the same periodicity, we establish $K$ parallel encoding branches per channel. Each branch is designed to extract a specific latent temporal view, creating a diverse pool of feature representations that allows the subsequent Channel-Adaptive Aggregation to selectively leverage the most relevant modeling expert for the unique dynamics of each channel~\cite{shazeer2017moe}. Within each branch, the forecasting horizon is mapped to a 2D latent grid of $G$ fixed anchor points. To further capture complex, non-standard local variations that transcend simple Gaussian shapes, at each anchor $g$, the model predicts parameters for $P$ Gaussian components. This hierarchical design---integrating diverse macroscopic views via $K$ and microscopic local details via $P$---enables \NAME~to robustly reconstruct intricate temporal structures across heterogeneous variables.

We adopt a mostly channel-independent strategy~\cite{patchtst, han2024capacity} where each variable is processed separately through identical network parameters, except for the Channel Adaptive Aggregation module, whose parameters are channel-specific to tailor the fusion of multi-branch features to the unique temporal dynamics of each channel. We denote $\mathbf{X}^{(i)} \in \mathbb{R}^{I \times 1}$ as the sequence of the $i$-th variable.

Additionally, we adopt the reversible instance normalization technique from RevIN~\cite{kim2021reversible} to preprocess inputs and postprocess outputs, enhancing robustness against temporal distribution variations.

\subsection{\EncodeBlockName}
To explicitly capture the complex coupling between intra-period and inter-period variations, we utilize the inherent periodicity commonly found in real-world data, such as daily cycles~\cite{lin2024cyclenet}. Instead of relying on frequency estimation methods like FFT~\cite{nussbaumer1981fft} which may be sensitive to noise, we define the period length $\mathcal{T}$ as a structural hyperparameter based on domain knowledge~\cite{lim2021tft}.

\textbf{Reshaping:} 
Given the historical time series $\mathbf{X}^{(i)} \in \mathbb{R}^{I \times 1}$, we reshape it into a 2D temporal tensor $\mathbf{X}_{2D}^{(i)}$ by folding the sequence based on $\mathcal{T}$. Specifically, the 1D sequence is organized into a tensor of shape $\mathbf{X}_{2D}^{(i)} \in \mathbb{R}^{N \times \mathcal{T} \times 1}$, where $\mathcal{T}$ represents the intra-period time steps and $N = \lceil I / \mathcal{T} \rceil$ represents the inter-period cycles. If $I$ is not perfectly divisible by $\mathcal{T}$, we apply zero-padding to the sequence to ensure alignment. After reshaping, the 2D tensor is upsampled to a fixed resolution $(N', \mathcal{T}')$ via bilinear interpolation, resulting in a tensor $\mathbf{X}_{up}^{(i)} \in \mathbb{R}^{N' \times \mathcal{T}' \times 1}$.
\begin{align}
    \mathbf{X}_{up}^{(i)} = \text{Upsample}(\text{Reshape}(\text{Padding}(\mathbf{X}^{(i)})))
\end{align}

\textbf{2D Feature Extraction:} 
To capture 2D temporal dynamics, we employ $K$ encoders, each functioning as a UNet-based Encoder. For the $k$-th branch, where $k=1, \dots, K$, the encoder maps the specific 2D temporal view $\mathbf{X}_{2D}$ to a latent representation that encodes the coupled temporal variations:
\begin{align}
    \mathbf{F}^{(i)}_{k} = \text{UNet}_k(\mathbf{X}_{up}^{(i)}),
\end{align}
where $\mathbf{F}^{(i)}_{k} \in \mathbb{R}^{N' \times \mathcal{T}' \times D_{c}}$ denotes the spatial feature map for the $k$-th branch.

To transform these spatial features into the input for the subsequent basis generator, we flatten the 2D feature map of each branch and project it via a corresponding linear layer:
\begin{align}
    \mathbf{Z}^{(i)}_k = \text{Linear}_k(\text{Flatten}(\mathbf{F}^{(i)}_{k})),
\end{align}
where $\mathbf{Z}^{(i)}_k \in \mathbb{R}^{G \times D_{h}}$ is the latent feature vector for the $k$-th branch, and $G$ represents the number of Gaussian kernels defined for the forecasting horizon.

\subsection{\DecodeBlockName}
Standard 2D Gaussian Splatting typically optimizes the position $\mu_k$ and the Cholesky factors of $\Sigma_k$ for free-floating Gaussian kernels. However, applying this directly to time series often leads to training instability. We observe that real-world time series exhibit pervasive structural recurrence, where intra-period and inter-period variations share underlying geometric templates despite stochastic fluctuations.
Guided by this insight and Dictionary Learning~\cite{aharon2006ksvd}, we fix the position of each Gaussian kernel and utilize a Fixed Basis Bank derived from discretized Cholesky parameters.

\textbf{Fixed Gaussian Basis Bank:} 
We pre-define a dictionary of $M$ Gaussian profiles $\mathcal{D} = \{B_1, \dots, B_M\}$. The covariance matrix is constructed as $\Sigma = \mathbf{L}\mathbf{L}^T$. 
Given a local offset $\delta=(\delta_x, \delta_y)^T$, the elliptically truncated Gaussian profile is defined as:
\begin{align}
    \hat{B}_m(\delta) 
    &= \exp \left( -\frac{1}{2} \delta^T \Sigma_m^{-1} \delta \right)
    \cdot \mathbb{I}_{\mathcal{E}_m}(\delta),
\end{align}
where $\mathbb{I}(\cdot)$ is the indicator function and $\mathcal{E}_m = \{\delta \mid \delta^T \Sigma_m^{-1}\delta \le 1\}$ denotes the elliptical support region induced by the covariance matrix of the $m$-th Gaussian basis.

Finally, each masked basis kernel is normalized over the discrete local support:
\begin{align}
    B_m(\delta) = \frac{\hat{B}_m(\delta)}{\sum_{\delta'} \hat{B}_m(\delta')}.
\end{align}
This fixed dictionary $\mathcal{D}$ remains frozen during training, transforming the unstable shape regression problem into a stable basis selection task.

\textbf{Basis Weight and Intensity Prediction:} 
We place $G$ Gaussian kernels at $G$ fixed positions. For each anchor, the model predicts how to combine the fixed bases to represent the local variation. Specifically, for the latent feature $\mathbf{Z}^{(i)}_k$, we employ a linear projection layer followed by a Softmax activation to generate the Basis Mixing Weights $\mathbf{W}^{(i)}_k \in \mathbb{R}^{G \times P \times M}$. This represents the probability distribution over the $M$ fixed basis kernels for each of the $P$ components. Simultaneously, an MLP projects $\mathbf{Z}^{(i)}_k$ to the Kernel Intensity $\mathbf{V}^{(i)}_k \in \mathbb{R}^{G \times P}$, which acts as a scalar amplitude controlling the strength of the variations.
\begin{align}
    \mathbf{W}^{(i)}_k &= \text{Softmax}( \text{Linear}_k (\mathbf{Z}^{(i)}_k)), \\
    \mathbf{V}^{(i)}_k &= \text{MLP}_k(\mathbf{Z}^{(i)}_k).
\end{align}

Consequently, the Composite Gaussian Kernel $\Phi^{(i)}_{k,g,p}$ at anchor $g$ of component $p$ in branch $k$ is dynamically constructed as:
\begin{align}
\Phi^{(i)}_{k,g,p}(\mathbf{\delta}) = \mathbf{V}^{(i)}_{k,g,p} \cdot \sum_{m=1}^{M} \mathbf{W}^{(i)}_{k, g, p, m} \cdot B_m(\delta),
\end{align}
This formulation effectively converts the difficult geometric regression problem into a stable dictionary learning.


\subsection{\RenderBlockName}

Our rendering strategy bridges the gap between 2D spatial representation and 1D temporal continuity.

\textbf{The Chronologically Discontinuity Problem:} 
A fundamental challenge in applying 2D vision techniques to time series lies in the topological mismatch between images and reshaped sequences. In images, the rightmost pixel of a row and the leftmost pixel of the subsequent row are spatially distant and semantically independent. In contrast, for a time series folded into a period-based 2D tensor, these two points are temporally adjacent ($t$ and $t+1$). In standard 2DGS, a Gaussian centered at $(x, y)$ influences neighbors based on 2D Euclidean distance. If applied to our reshaped series, a Gaussian located at the far right would fail to influence its immediate temporal successor at the start of the next row, creating a break in the forecast.




To enforce temporal continuity and effectively model periodic patterns, we propose a \RenderBlockName block, which implements the rasterization not as a spatial overlay, but as a continuous temporal overlay.



\begin{figure}[!t]
    \centering
    \includegraphics[width=\linewidth]{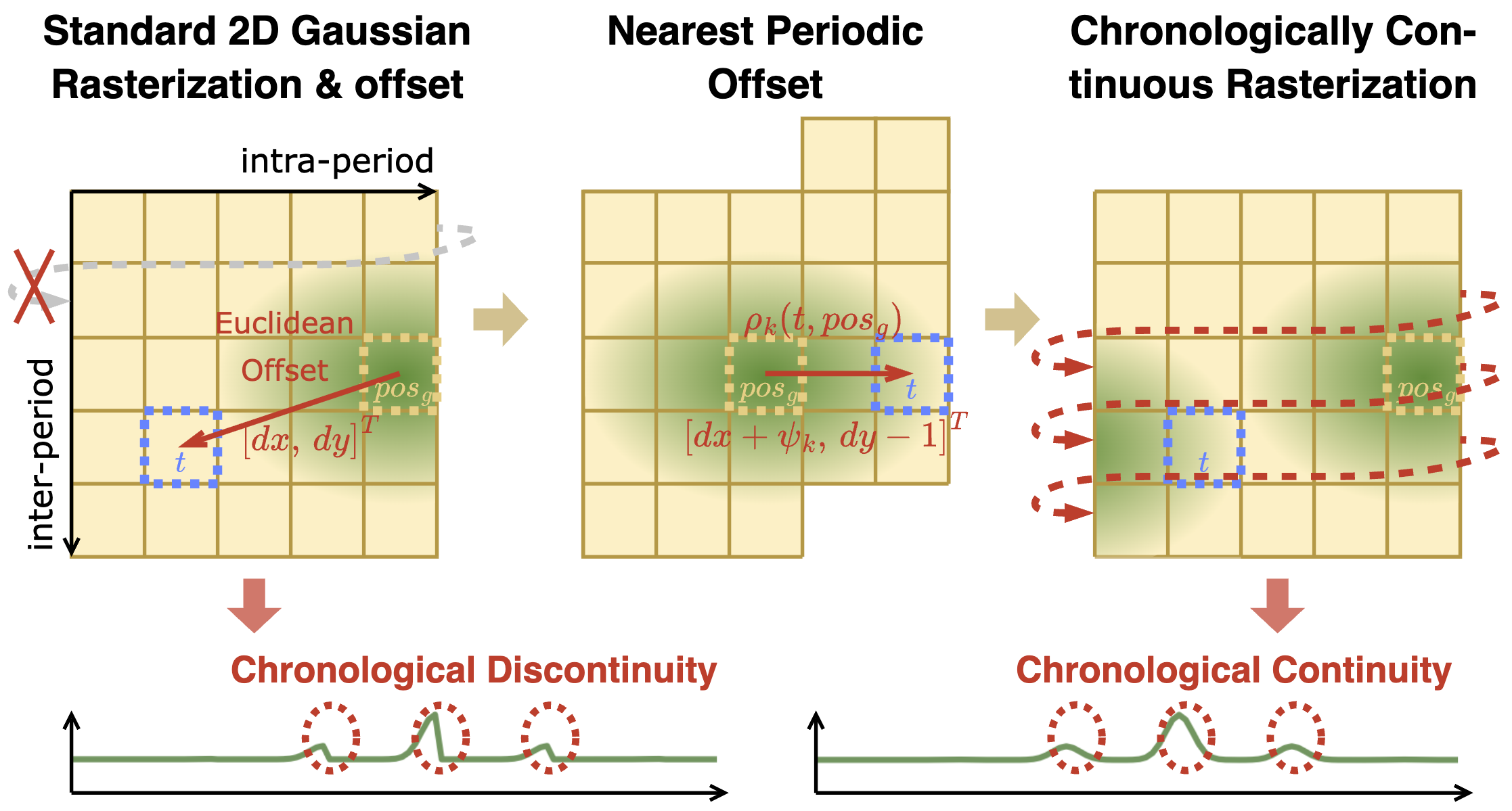}
    \caption{
    Illustration of chronological discontinuity in folded time-series grids. Standard 2D rasterization breaks kernel influence across period boundaries, whereas our chronologically continuous rasterization preserves temporal continuity by following the unfolded temporal order.
    }
    \vspace{-3pt}
    \label{fig:render}
\end{figure}

\textbf{Theoretical Formulation:}
We define the temporal dynamics of branch $k$ on a 2D space $\mathcal{S}_k$, whose width is fixed to $\psi_k$. 
Here, $\psi_k$ denotes the period length used by branch $k$; the horizontal dimension represents the phase within a period, while the vertical dimension indexes different periods. 

For a temporal index $t$ and a Gaussian center $\text{pos}_g$, the rendering process is formulated as:
\begin{align}
\mathbf{R}^{(i)}_{k,:,t} = \sum_{g} \Phi^{(i)}_{k, g, :}(\rho_k(t, \text{pos}_g)),
\end{align}
where $\mathbf{R}^{(i)}_k$ is the rendered result from branch $k$, and $\rho_k(t,\text{pos}_g)$ denotes the nearest periodic offset between the 2D coordinates obtained by mapping the one-dimensional indices $t$ and $\text{pos}_g$ into $\mathcal{S}_k$.
Since the temporal index starts from 1, the horizontal and vertical offsets are computed as
$dx = ((t-1) \bmod \psi_k) - ((\text{pos}_g-1) \bmod \psi_k)$ and
$dy = \lfloor (t-1)/\psi_k \rfloor - \lfloor (\text{pos}_g-1)/\psi_k \rfloor$.

Considering the periodic structure along the horizontal dimension, the nearest periodic offset is defined as:
\begin{align}
\rho_k(t, \text{pos}_g) =
\begin{cases}
\left[ dx,\, dy \right]^T, 
& |dx| \le \psi_k/2, \\[2mm]
\left[ dx-\psi_k,\, dy+1 \right]^T, 
& dx > \psi_k/2, \\[2mm]
\left[ dx+\psi_k,\, dy-1 \right]^T, 
& dx < -\psi_k/2 .
\end{cases}
\end{align}
This operation evaluates the kernel using the nearest horizontal periodic copy of its center, with the vertical offset adjusted to account for the corresponding period shift. 
Thus, the rendering process can model both intraperiod dependencies and interperiod consistency.

\textbf{Practical Implementation:}
Since operating on an infinite 2D plane is computationally infeasible, we implement the mechanism through a sequence of discrete operations that approximate the theoretical model. 
We rasterize the learned Gaussian kernel $\Phi^{(i)}_{k,g,p}(\mathbf{\delta})$ onto a 2D tensor $\mathcal{G}^{(i)}_{k,g,p} \in \mathbb{R}^{h \times w}$.
Then, we address the column dimension constraint. 
We constrain the local kernel width $w$ to be no larger than the branch period $\psi_k$. 
We symmetrically pad the columns of the kernel with zeros until the total number of columns equals $\psi_k$. We then flatten the padded 2D kernel sequence into a 1D sequence. Finally, we place the flattened kernel onto the target temporal canvas. The placement is aligned such that the center of the flattened kernel corresponds to $\text{pos}_g$ in the 1D time series.
\begin{align}
\tilde{Y}^{(i)}_k = \sum_{g} \text{Shift}\left(\text{Flatten}\left(\text{Padding}\left(\mathcal{G}^{(i)}_{k, g, :}\right)\right), \text{pos}_g\right),
\end{align}
where $\tilde{Y}^{(i)}_k \in \mathbb{R}^{P \times O}$ is the rendered result from branch $k$, and $\text{Shift}(\mathbf{v}, c)$ denotes placing the 1D sequence $\mathbf{v}$ such that its center aligns with position $c$ on the output canvas. Here, $\text{pos}_g$ directly denotes the temporal coordinate of the $g$-th Gaussian center on the output canvas.

\subsection{\AggregateBlockName}
The final step is derived by aggregating multiple forecasts.
Since different channels in a multivariate dataset may exhibit different dominant periods and different channels may exhibit different temporal variation characteristics in the same period, simply averaging the branches is suboptimal.
We introduce two sets of learnable parameters for each channel $i$:
(1) Branch Weights $\Gamma \in \mathbb{R}^{C \times K}$: Capturing the importance of each branch for channel $i$.
(2) Component Weights $\Omega \in \mathbb{R}^{C \times P}$: Capturing the importance of each component group $p$ for channel $i$.

We apply Softmax normalization to ensure stability and the final prediction $\hat{\mathbf{Y}}^{(i)} \in \mathbb{R}^O$ for channel $i$ is the weighted sum of all rendered candidates:
\begin{align}
\gamma_{i,k} &= \frac{\exp(\Gamma_{i,k})}{\sum_{j} \exp(\Gamma_{i,j})}, \\
\omega_{i,p} &= \frac{\exp(\Omega_{i,p})}{\sum_{j} \exp(\Omega_{i,j})}, \\
\hat{\mathbf{Y}}^{(i)} &= \sum_{k=1}^{K} \sum_{p=1}^{P} \gamma_{i,k} \cdot \omega_{i,p} \cdot \tilde{Y}_{k,p,:}^{(i)}.
\end{align}

This Adaptive Fusion mechanism allows \NAME to automatically select the most appropriate temporal views and feature components for every individual variable, significantly enhancing modeling flexibility.


\section{Experiments}

\renewcommand{\arraystretch}{0.8}
\begin{table*}[!htbp]
\caption{Full results of the long-term forecasting task. The input sequence length is set to 96 for \NAME and all Baselines. ``Avg'' refers to the average results from all four prediction results. The best results are highlighted in \textbf{bold}, and the second-best results are \underline{underlined}.}
\label{tabs:main_results}
\resizebox{1\linewidth}{!}{
\begin{tabular}{cc|cc|cc|cc|cc|cc|cc|cc|cc|cc|cc|cc}
\toprule
\multicolumn{2}{c|}{\multirow{2}{*}{Model}} & \multicolumn{2}{c|}{\textbf{\NAME}} & \multicolumn{2}{c|}{WPMixer} & \multicolumn{2}{c|}{TimeMixer} & \multicolumn{2}{c|}{iTransformer} & \multicolumn{2}{c|}{PatchTST} & \multicolumn{2}{c|}{TimesNet} & \multicolumn{2}{c|}{Crossformer} & \multicolumn{2}{c|}{MICN} & \multicolumn{2}{c|}{DLinear} & \multicolumn{2}{c|}{FEDformer} & \multicolumn{2}{c}{Autoformer} \\
\multicolumn{2}{c|}{} & \multicolumn{2}{c|}{\textbf{(Ours)}} & \multicolumn{2}{c|}{(2025)} & \multicolumn{2}{c|}{(2024)} & \multicolumn{2}{c|}{(2024)} & \multicolumn{2}{c|}{(2023)} & \multicolumn{2}{c|}{(2023)} & \multicolumn{2}{c|}{(2023)} & \multicolumn{2}{c|}{(2023)} & \multicolumn{2}{c|}{(2023)} & \multicolumn{2}{c|}{(2022)} & \multicolumn{2}{c}{(2021)} \\
\multicolumn{2}{c|}{Metric} & MSE & MAE & MSE & MAE & MSE & MAE & MSE & MAE & MSE & MAE & MSE & MAE & MSE & MAE & MSE & MAE & MSE & MAE & MSE & MAE & MSE & MAE \\ \midrule
\multicolumn{1}{c|}{\multirow{5}{*}{\rotatebox{90}{Weather}}} & 96 & \textbf{0.161} & \textbf{0.199} & 0.165 & \underline{0.207} & \underline{0.163} & 0.209 & 0.174 & 0.214 & 0.186 & 0.227 & 0.172 & 0.220 & 0.195 & 0.271 & 0.198 & 0.261 & 0.195 & 0.252 & 0.217 & 0.296 & 0.266 & 0.336 \\
\multicolumn{1}{c|}{} & 192 & \textbf{0.206} & \textbf{0.242} & 0.211 & \underline{0.249} & \underline{0.208} & 0.250 & 0.221 & 0.254 & 0.234 & 0.265 & 0.219 & 0.261 & 0.209 & 0.277 & 0.239 & 0.299 & 0.237 & 0.295 & 0.276 & 0.336 & 0.307 & 0.367 \\
\multicolumn{1}{c|}{} & 336 & 0.263 & \textbf{0.286} & 0.265 & 0.289 & \underline{0.251} & \underline{0.287} & 0.278 & 0.296 & 0.284 & 0.301 & \textbf{0.246} & 0.337 & 0.273 & 0.332 & 0.285 & 0.336 & 0.282 & 0.331 & 0.339 & 0.380 & 0.359 & 0.395 \\
\multicolumn{1}{c|}{} & 720 & \underline{0.344} & \textbf{0.337} & \textbf{0.339} & \underline{0.340} & \textbf{0.339} & 0.341 & 0.358 & 0.347 & 0.356 & 0.349 & 0.365 & 0.359 & 0.379 & 0.401 & 0.351 & 0.388 & 0.345 & 0.382 & 0.403 & 0.428 & 0.419 & 0.428 \\

\cmidrule(lr){2-24}

\multicolumn{1}{c|}{} & Avg & \underline{0.244} & \textbf{0.266} & 0.245 & 
\underline{0.271} & \textbf{0.240} & \underline{0.271} & 0.258 & 0.278 & 0.265 & 0.285 & 0.251 & 0.294 & 0.264 & 0.320 & 0.268 & 0.321 & 0.265 & 0.315 & 0.309 & 0.360 & 0.338 & 0.382 \\ \midrule
\multicolumn{1}{c|}{\multirow{5}{*}{\rotatebox{90}{Electricity}}} & 96 & 0.151 & \textbf{0.237} & \underline{0.150} & 0.241 & 0.153 & 0.247 & \textbf{0.148} & \underline{0.240} & 0.190 & 0.296 & 0.168 & 0.272 & 0.219 & 0.314 & 0.180 & 0.293 & 0.210 & 0.302 & 0.193 & 0.308 & 0.201 & 0.317 \\
\multicolumn{1}{c|}{} & 192 & \underline{0.164} & \textbf{0.250} & \textbf{0.162} & \underline{0.253} & 0.166 & 0.256 & \textbf{0.162} & \underline{0.253} & 0.199 & 0.304 & 0.184 & 0.322 & 0.231 & 0.322 & 0.189 & 0.302 & 0.210 & 0.305 & 0.201 & 0.315 & 0.222 & 0.334 \\
\multicolumn{1}{c|}{} & 336 & 0.184 & \underline{0.270} & \underline{0.181} & 0.272 & 0.185 & 0.277 & \textbf{0.178} & \textbf{0.269} & 0.217 & 0.319 & 0.198 & 0.300 & 0.246 & 0.337 & 0.198 & 0.312 & 0.223 & 0.319 & 0.214 & 0.329 & 0.231 & 0.443 \\
\multicolumn{1}{c|}{} & 720 & 0.224 & \textbf{0.304} & \underline{0.219} & \underline{0.306} & 0.225 & 0.310 & 0.225 & 0.317 & 0.258 & 0.352 & 0.220 & 0.320 & 0.280 & 0.363 & \textbf{0.217} & 0.330 & 0.258 & 0.350 & 0.246 & 0.355 & 0.254 & 0.361 \\

\cmidrule(lr){2-24}

\multicolumn{1}{c|}{} & Avg & \underline{0.181} & \textbf{0.265} & \textbf{0.178} & \underline{0.268} & 0.182 & 0.272 & \textbf{0.178} & 0.270 & 0.216 & 0.318 & 0.193 & 0.304 & 0.244 & 0.334 & 0.196 & 0.309 & 0.225 & 0.319 & 0.214 & 0.327 & 0.227 & 0.338 \\ \midrule

\multicolumn{1}{c|}{\multirow{5}{*}{\rotatebox{90}{Traffic}}} & 96 & \underline{0.457} & \underline{0.283} & 0.467 & 0.286 & 0.462 & 0.285 & \textbf{0.395} & \textbf{0.268} & 0.526 & 0.347 & 0.593 & 0.321 & 0.644 & 0.429 & 0.577 & 0.350 & 0.650 & 0.396 & 0.587 & 0.366 & 0.613 & 0.388 \\
\multicolumn{1}{c|}{} & 192 & \underline{0.463} & \underline{0.282} & 0.471 & 0.288 & 0.473 & 0.296 & \textbf{0.417} & \textbf{0.276} & 0.522 & 0.332 & 0.617 & 0.336 & 0.665 & 0.431 & 0.589 & 0.356 & 0.598 & 0.370 & 0.604 & 0.373 & 0.616 & 0.382 \\
\multicolumn{1}{c|}{} & 336 & \underline{0.478} & \underline{0.288} & 0.487 & 0.297 & 0.498 & 0.296 & \textbf{0.433} & \textbf{0.283} & 0.517 & 0.334 & 0.629 & 0.336 & 0.674 & 0.420 & 0.594 & 0.358 & 0.605 & 0.373 & 0.621 & 0.383 & 0.622 & 0.337 \\
\multicolumn{1}{c|}{} & 720 & 0.514 & \underline{0.310} & 0.526 & 0.317 & \underline{0.506} & 0.313 & \textbf{0.467} & \textbf{0.302} & 0.552 & 0.352 & 0.640 & 0.350 & 0.683 & 0.424 & 0.613 & 0.361 & 0.645 & 0.394 & 0.626 & 0.382 & 0.660 & 0.408 \\

\cmidrule(lr){2-24}

\multicolumn{1}{c|}{} & Avg & \underline{0.478} & \underline{0.291} & 0.488 & 0.297 & 0.484 & 0.297 & \textbf{0.428} & \textbf{0.282} & 0.529 & 0.341 & 0.620 & 0.336 & 0.667 & 0.426 & 0.593 & 0.356 & 0.625 & 0.383 & 0.610 & 0.376 & 0.628 & 0.379 \\ \midrule

\multicolumn{1}{c|}{\multirow{5}{*}{\rotatebox{90}{ETTh1}}} & 96 & \textbf{0.361} & \textbf{0.386} & \underline{0.371} & \underline{0.393} & 0.375 & 0.400 & 0.386 & 0.405 & 0.460 & 0.447 & 0.384 & 0.402 & 0.423 & 0.448 & 0.426 & 0.446 & 0.397 & 0.412 & 0.395 & 0.424 & 0.449 & 0.459 \\
\multicolumn{1}{c|}{} & 192 & \textbf{0.410} & \textbf{0.419} & \underline{0.428} & \underline{0.421} & 0.429 & \underline{0.421} & 0.441 & 0.436 & 0.512 & 0.477 & 0.436 & 0.429 & 0.471 & 0.474 & 0.454 & 0.464 & 0.446 & 0.441 & 0.469 & 0.470 & 0.500 & 0.482 \\
\multicolumn{1}{c|}{} & 336 & \textbf{0.452} & \underline{0.440} & \underline{0.473} & \textbf{0.435} & 0.484 & 0.458 & 0.487 & 0.458 & 0.546 & 0.496 & 0.638 & 0.469 & 0.570 & 0.546 & 0.493 & 0.487 & 0.489 & 0.467 & 0.530 & 0.499 & 0.521 & 0.496 \\
\multicolumn{1}{c|}{} & 720 & \textbf{0.451} & \underline{0.451} & \underline{0.453} & \textbf{0.448} & 0.498 & 0.482 & 0.503 & 0.491 & 0.544 & 0.517 & 0.521 & 0.500 & 0.653 & 0.621 & 0.526 & 0.526 & 0.513 & 0.510 & 0.598 & 0.544 & 0.514 & 0.512 \\

\cmidrule(lr){2-24}

\multicolumn{1}{c|}{} & Avg & \textbf{0.419} & \textbf{0.424} & \underline{0.431} & \textbf{0.424} & 0.447 & \underline{0.440} & 0.454 & 0.447 & 0.516 & 0.484 & 0.495 & 0.450 & 0.529 & 0.522 & 0.475 & 0.480 & 0.461 & 0.457 & 0.498 & 0.484 & 0.496 & 0.487 \\ \midrule

\multicolumn{1}{c|}{\multirow{5}{*}{\rotatebox{90}{ETTh2}}} & 96 & \textbf{0.281} & \textbf{0.329} & \underline{0.286} & \underline{0.337} & 0.289 & 0.341 & 0.297 & 0.349 & 0.308 & 0.355 & 0.340 & 0.374 & 0.745 & 0.584 & 0.372 & 0.424 & 0.340 & 0.394 & 0.358 & 0.397 & 0.346 & 0.388 \\
\multicolumn{1}{c|}{} & 192 & \textbf{0.359} & \textbf{0.380} & \underline{0.363} & \underline{0.391} & 0.372 & 0.392 & 0.380 & 0.400 & 0.393 & 0.405 & 0.402 & 0.414 & 0.877 & 0.656 & 0.492 & 0.492 & 0.482 & 0.479 & 0.429 & 0.439 & 0.456 & 0.452 \\
\multicolumn{1}{c|}{} & 336 & 0.401 & 0.416 & \textbf{0.378} & \textbf{0.407} & \underline{0.386} & \underline{0.414} & 0.428 & 0.432 & 0.427 & 0.436 & 0.452 & 0.452 & 1.043 & 0.731 & 0.607 & 0.555 & 0.591 & 0.541 & 0.496 & 0.487 & 0.482 & 0.486 \\
\multicolumn{1}{c|}{} & 720 & \textbf{0.410} & \textbf{0.431} & 0.419 & 0.437 & \underline{0.412} & \underline{0.434} & 0.427 & 0.445 & 0.436 & 0.450 & 0.462 & 0.468 & 1.104 & 0.763 & 0.824 & 0.655 & 0.839 & 0.661 & 0.463 & 0.474 & 0.515 & 0.511 \\

\cmidrule(lr){2-24}

\multicolumn{1}{c|}{} & Avg & \underline{0.363} & \textbf{0.389} & \textbf{0.362} & \underline{0.393} & 0.364 & 0.395 & 0.383 & 0.407 & 0.391 & 0.411 & 0.414 & 0.427 & 0.942 & 0.684 & 0.574 & 0.531 & 0.563 & 0.519 & 0.437 & 0.449 & 0.450 & 0.459 \\ \midrule

\multicolumn{1}{c|}{\multirow{5}{*}{\rotatebox{90}{ETTm1}}} & 96 & \textbf{0.307} & \textbf{0.347} & \underline{0.317} & \underline{0.352} & 0.320 & 0.357 & 0.334 & 0.368 & 0.352 & 0.374 & 0.338 & 0.375 & 0.404 & 0.426 & 0.365 & 0.387 & 0.346 & 0.374 & 0.379 & 0.419 & 0.505 & 0.475 \\
\multicolumn{1}{c|}{} & 192 & \textbf{0.354} & \textbf{0.373} & \underline{0.359} & \underline{0.376} & 0.361 & 0.381 & 0.377 & 0.391 & 0.390 & 0.393 & 0.374 & 0.387 & 0.450 & 0.451 & 0.403 & 0.408 & 0.382 & 0.391 & 0.426 & 0.441 & 0.553 & 0.496 \\
\multicolumn{1}{c|}{} & 336 & \textbf{0.383} & \textbf{0.397} & \underline{0.387} & \underline{0.399} & 0.390 & 0.404 & 0.426 & 0.420 & 0.421 & 0.414 & 0.410 & 0.411 & 0.532 & 0.515 & 0.436 & 0.431 & 0.415 & 0.415 & 0.445 & 0.459 & 0.621 & 0.537 \\
\multicolumn{1}{c|}{} & 720 & \textbf{0.443} & \textbf{0.433} & \underline{0.449} & \underline{0.435} & 0.454 & 0.441 & 0.491 & 0.459 & 0.462 & 0.449 & 0.478 & 0.450 & 0.666 & 0.589 & 0.489 & 0.462 & 0.473 & 0.451 & 0.543 & 0.490 & 0.671 & 0.561 \\

\cmidrule(lr){2-24}

\multicolumn{1}{c|}{} & Avg & \textbf{0.372} & \textbf{0.388} & \underline{0.378} & \underline{0.391} & 0.381 & 0.395 & 0.407 & 0.410 & 0.406 & 0.407 & 0.400 & 0.406 & 0.513 & 0.495 & 0.423 & 0.422 & 0.404 & 0.408 & 0.448 & 0.452 & 0.588 & 0.517 \\ \midrule

\multicolumn{1}{c|}{\multirow{5}{*}{\rotatebox{90}{ETTm2}}} & 96 & \textbf{0.169} & \textbf{0.251} & \underline{0.172} & \underline{0.253} & 0.175 & 0.258 & 0.180 & 0.264 & 0.183 & 0.270 & 0.187 & 0.267 & 0.287 & 0.366 & 0.197 & 0.296 & 0.193 & 0.293 & 0.203 & 0.287 & 0.255 & 0.339 \\
\multicolumn{1}{c|}{} & 192 & 0.239 & \textbf{0.293} & \textbf{0.234} & \underline{0.295} & \underline{0.237} & 0.299 & 0.250 & 0.309 & 0.255 & 0.314 & 0.249 & 0.309 & 0.414 & 0.492 & 0.284 & 0.361 & 0.284 & 0.361 & 0.269 & 0.328 & 0.281 & 0.340 \\
\multicolumn{1}{c|}{} & 336 & \underline{0.298} & \textbf{0.332} & \textbf{0.292} & \underline{0.334} & \underline{0.298} & 0.340 & 0.311 & 0.348 & 0.309 & 0.347 & 0.321 & 0.351 & 0.597 & 0.542 & 0.381 & 0.429 & 0.382 & 0.429 & 0.325 & 0.366 & 0.339 & 0.372 \\
\multicolumn{1}{c|}{} & 720 & 0.396 & \textbf{0.391} & \textbf{0.389} & \underline{0.392} & \underline{0.391} & 0.396 & 0.412 & 0.407 & 0.412 & 0.404 & 0.408 & 0.403 & 1.730 & 1.042 & 0.549 & 0.522 & 0.558 & 0.525 & 0.421 & 0.415 & 0.433 & 0.432 \\

\cmidrule(lr){2-24}

\multicolumn{1}{c|}{} & Avg & 0.276 & \textbf{0.317} & \textbf{0.272} & \underline{0.319} & \underline{0.275} & 0.323 & 0.288 & 0.332 & 0.290 & 0.334 & 0.291 & 0.333 & 0.757 & 0.610 & 0.353 & 0.402 & 0.354 & 0.402 & 0.305 & 0.349 & 0.327 & 0.371 \\ 
\bottomrule
\end{tabular}}
\end{table*}

\renewcommand{\arraystretch}{1}

To evaluate the performance of our proposed \NAME, we conduct extensive experiments on 7 widely used real-world benchmarks. We compare our model against state-of-the-art baselines under a unified experimental setting.

\subsection{Experimental Settings}

\textbf{Datasets:}
We utilize 7 standard datasets covering various domains, including energy, traffic, and weather~\cite{wang2024benchmark}. 

\textbf{Baselines:}
We compare \NAME with extensive state-of-the-art models, categorized into three groups:

(1) Transformer-based methods, which includes iTransformer ~\cite{liuitransformer}, PatchTST ~\cite{patchtst}, Crossformer ~\cite{zhang2023crossformer}, FEDformer ~\cite{zhou2022fedformer}, Autoformer ~\cite{wu2021autoformer}. These models represent the mainstream approaches focusing on global dependency modeling.

(2) MLP-based methods, which includes TimeMixer ~\cite{wang2024timemixer}, WPMixer ~\cite{murad2025wpmixer} and DLinear ~\cite{dlinear}. These models utilize simple linear mappings with decomposition or multi-scale mixing strategies.

(3) CNN-based methods, which includes TimesNet ~\cite{wu2022timesnet} and MICN ~\cite{wang2023micn}. These methods are the most relevant competitors as they also explore 2D temporal variations.

\textbf{Unified Experiment Settings:}
Following the previous works, we adopt a unified experimental setup to ensure a fair comparison.
The input look-back window length is fixed at $I=96$ for all datasets. The prediction horizons are set to $O \in \{96, 192, 336, 720\}$.
We employ Mean Squared Error (MSE) and Mean Absolute Error (MAE) as the evaluation metrics. Lower MSE and MAE indicate better prediction accuracy.

\subsection{Main Results}

The forecasting results are presented in Table \ref{tabs:main_results}. 
Overall, \NAME achieves competitive performance across different datasets and prediction horizons, and obtains the best or second-best results in many cases. 
In particular, on datasets such as Weather, ETTh1, and ETTm1, \NAME shows consistent improvements or comparable results over strong baselines. 
These results suggest that the proposed generative rendering paradigm can effectively model temporal patterns after 2D reshaping, especially when both intra-period and inter-period variations are involved.

Compared with 1D sequence modeling methods such as iTransformer and TimeMixer, \NAME often achieves lower forecasting errors, indicating that introducing a 2D temporal representation can provide useful structural information for long-term forecasting. 
Compared with TimesNet, which also constructs 2D temporal representations, \NAME obtains better average performance on several benchmarks, suggesting that the 2D Gaussian Splatting module contributes additional modeling capacity beyond the reshaping operation itself.

\subsection{Model Analysis}
To verify the effectiveness of each component of our proposed \NAME, we conduct extensive ablation studies. We systematically remove or replace key modules to analyze their contribution to the final performance.  

\textbf{Effectiveness of \EncodeBlockName block:}
We investigated the necessity of reshaping the sequence and using a strong visual backbone, namely the UNet-based encoder, to extract coupled temporal features. We replace the \EncodeBlockNameShort with three variants: 
(1) Linear: Without reshaping the input, we replace the 2D CNN backbone with a simple linear layer. 
(2) MLP: Without reshaping the input, we replace the 2D CNN backbone with a Multi-Layer Perceptron. 
(3) Plain CNN: Using a standard stack of convolutional layers without the U-shape skip connections and multi-scale design of the UNet.
As shown in Table \ref{tabs:ablation_enc}, the performance drops significantly when replacing the 2D backbone with Linear or MLP variants. This substantial gap highlights the limitations of processing time series as flat 1D sequences. Without reshaping, simple 1D projections struggle to explicitly disentangle and model the complex coupling between intra-period dynamics and inter-period evolution, leading to suboptimal feature representation. While the Plain CNN variant introduces 2D processing, it still lags behind our UNet-based design. This suggests that a simple stack of convolutional layers is insufficient for capturing the intricate 2D temporal variations. The UNet architecture, characterized by its U-shape structure and skip connections, proves crucial. It allows the model to simultaneously aggregate global context while preserving local details, both of which are essential for the precise generation of Gaussian kernels.
\renewcommand{\arraystretch}{0.9}
\begin{table}[htb]
\caption{Ablations on the \EncodeBlockName.}
\label{tabs:ablation_enc}

\resizebox{0.48\textwidth}{!}
{
\begin{tabular}{c|cc|cc|cc|cc}

\toprule

\multicolumn{1}{c|}{\multirow{2}{*}{Method}} & \multicolumn{2}{c|}{\multirow{1}{*}{ETTh1}} & \multicolumn{2}{c|}{\multirow{1}{*}{ETTm1}} & \multicolumn{2}{c|}{\multirow{1}{*}{Weather}} & \multicolumn{2}{c}{\multirow{1}{*}{Electricity}} \\


\cmidrule(lr){2-3} \cmidrule(lr){4-5} \cmidrule(lr){6-7} \cmidrule(lr){8-9}

  & MSE & MAE & MSE & MAE & MSE & MAE & MSE & MAE \\

\toprule
Origin & \textbf{0.419} & \textbf{0.424} & \textbf{0.372} & \textbf{0.388} & 
         \textbf{0.244} & \textbf{0.266} & \textbf{0.181} & \textbf{0.265} \\

\midrule
Linear & 0.454 & 0.439 & 0.390 & 0.395 & 0.252 & 0.275 & 0.192 & 0.275 \\

\midrule
MLP & 0.441 & 0.430 & 0.390 & 0.396 & 0.249 & 0.271 & 0.186 & 0.271 \\

\midrule
Plain CNN & 0.442 & 0.431 & 0.390 & 0.395 & 0.251 & 0.274 & 0.190 & 0.274 \\

\bottomrule

\end{tabular}
}

\end{table}
\renewcommand{\arraystretch}{1}

\textbf{Effectiveness of \DecodeBlockName block:}
We investigated the necessity of utilizing the multi-basis mechanism. We compare it against w/o Multi-Basis: We remove the fixed Gaussian basis bank and directly regress the Cholesky factors from the features.
As shown in Table \ref{tabs:ablation_dec}, the removal of the Multi-Basis mechanism results in a catastrophic performance degradation across all benchmarks. Most notably, on the Electricity dataset, the MSE surges dramatically from 0.181 to 0.877, and the MAE surges dramatically from 0.265 to 0.767. This dramatic collapse serves as strong empirical evidence supporting one of our core motivations: directly regressing geometric parameters (Cholesky factors) is highly unstable for time series forecasting. By transforming the difficult shape regression problem into a stable weight prediction task over a pre-defined dictionary, the \DecodeBlockNameShort ensures that the generated patterns remain robust and plausible.
\begin{table}[htb]
\caption{Ablations on the \DecodeBlockName.}
\label{tabs:ablation_dec}

\resizebox{0.48\textwidth}{!}
{
\begin{tabular}{c|cc|cc|cc|cc}

\toprule

\multicolumn{1}{c|}{\multirow{2}{*}{Method}} & \multicolumn{2}{c|}{\multirow{1}{*}{ETTh1}} & \multicolumn{2}{c|}{\multirow{1}{*}{ETTm1}} & \multicolumn{2}{c|}{\multirow{1}{*}{Weather}} & \multicolumn{2}{c}{\multirow{1}{*}{Electricity}} \\


\cmidrule(lr){2-3} \cmidrule(lr){4-5} \cmidrule(lr){6-7} \cmidrule(lr){8-9}

  & MSE & MAE & MSE & MAE & MSE & MAE & MSE & MAE \\

\toprule
Origin & \textbf{0.419} & \textbf{0.424} & \textbf{0.372} & \textbf{0.388} & 
         \textbf{0.244} & \textbf{0.266} & \textbf{0.181} & \textbf{0.265} \\

\midrule
w/o Multi-Basis & 0.552 & 0.500 & 0.695 & 0.552 & 0.281 & 0.297 & 0.877 & 0.767 \\

\bottomrule

\end{tabular}
}

\end{table}

\textbf{Effectiveness of Rendering paradigm:}
To validate the fundamental advantage of our generative rendering paradigm over traditional regression, we replace the Gaussian generation and rasterization blocks with direct feature mapping heads:
(1) Linear: Projects encoder features via a single linear layer.
(2) MLP: Maps encoder features using a Multi-Layer Perceptron.
As shown in Table \ref{tabs:ablation_ren}, the proposed generative rendering paradigm yields superior or highly competitive performance compared to traditional regression baselines in general. The Linear variant performs the worst in most cases, which confirms that a simple linear mapping is insufficient to decode the complex high-dimensional features extracted by the encoder for forecasting. While the MLP variant improves upon the Linear baseline by introducing non-linearity, it still generally lags behind our rendering-based approach. Our Gaussian Rendering mechanism, by constructing time series from Gaussian kernels, is better equipped to preserve the structural integrity of the time series.
\begin{table}[htb]
\caption{Ablations on the Rendering.}
\label{tabs:ablation_ren}

\resizebox{0.48\textwidth}{!}
{
\begin{tabular}{c|cc|cc|cc|cc}

\toprule

\multicolumn{1}{c|}{\multirow{2}{*}{Method}} & \multicolumn{2}{c|}{\multirow{1}{*}{ETTh1}} & \multicolumn{2}{c|}{\multirow{1}{*}{ETTm1}} & \multicolumn{2}{c|}{\multirow{1}{*}{Weather}} & \multicolumn{2}{c}{\multirow{1}{*}{Electricity}} \\


\cmidrule(lr){2-3} \cmidrule(lr){4-5} \cmidrule(lr){6-7} \cmidrule(lr){8-9}

 & MSE & MAE & MSE & MAE & MSE & MAE & MSE & MAE \\

\toprule
Origin & \textbf{0.419} & \textbf{0.424} & \textbf{0.372} & \textbf{0.388} & 
         \textbf{0.244} & \textbf{0.266} & 0.181 & \textbf{0.265} \\

\midrule
Linear & 0.440 & 0.435 & 0.379 & 0.396 & 0.245 & 0.268 & 0.182 & 0.267 \\

\midrule
MLP & 0.423 & 0.427 & 0.374 & 0.391 & 0.245 & 0.268 & \textbf{0.179} & \textbf{0.265} \\

\bottomrule

\end{tabular}
}

\end{table}

\textbf{Effectiveness of \AggregateBlockName block:}
We evaluate the channel-adaptive weighting mechanism, which fuses multiple forecasts, by comparing it with two variants:
(1) Average: Replacing the learnable weights with a simple average over all branches and all components.
(2) Channel Agnostic: Using learnable weights that are shared across all variates.
As shown in Table \ref{tabs:ablation_agg}, our \AggregateBlockNameShort consistently yields the better results, validating the necessity of channel-wise weighting strategies. The Average variant performs worse than the learnable methods, indicating that different temporal views and Gaussian components contribute unequally to the final forecast. On datasets with diverse variable characteristics, such as Weather, the Channel Agnostic approach lags significantly behind our \AggregateBlockNameShort. This suggests that sharing a single set of weights across all variates is insufficient to model the unique temporal dynamics of each channel.
\begin{table}[htb]
\caption{Ablations on the \AggregateBlockName.}
\label{tabs:ablation_agg}

\resizebox{0.48\textwidth}{!}
{
\begin{tabular}{c|cc|cc|cc|cc}

\toprule

\multicolumn{1}{c|}{\multirow{2}{*}{Method}} & \multicolumn{2}{c|}{\multirow{1}{*}{ETTh1}} & \multicolumn{2}{c|}{\multirow{1}{*}{ETTm1}} & \multicolumn{2}{c|}{\multirow{1}{*}{Weather}} & \multicolumn{2}{c}{\multirow{1}{*}{Electricity}} \\


\cmidrule(lr){2-3} \cmidrule(lr){4-5} \cmidrule(lr){6-7} \cmidrule(lr){8-9}

 & MSE & MAE & MSE & MAE & MSE & MAE & MSE & MAE \\

\toprule
Origin & \textbf{0.419} & \textbf{0.424} & \textbf{0.372} & \textbf{0.388} & 
         \textbf{0.244} & \textbf{0.266} & \textbf{0.181} & \textbf{0.265} \\

\midrule
Average & 0.428 & 0.430 & 0.378 & 0.391 & 0.252 & 0.272 & 0.191 & 0.272 \\

\midrule
Channel Agnostic & \textbf{0.419} & \textbf{0.424} & 0.376 & 0.389 & 0.253 & 0.271 & 0.190 & 0.271 \\

\bottomrule

\end{tabular}
}

\end{table}

\begin{figure}[t!]
  \centering
    \includegraphics[width=1\linewidth]{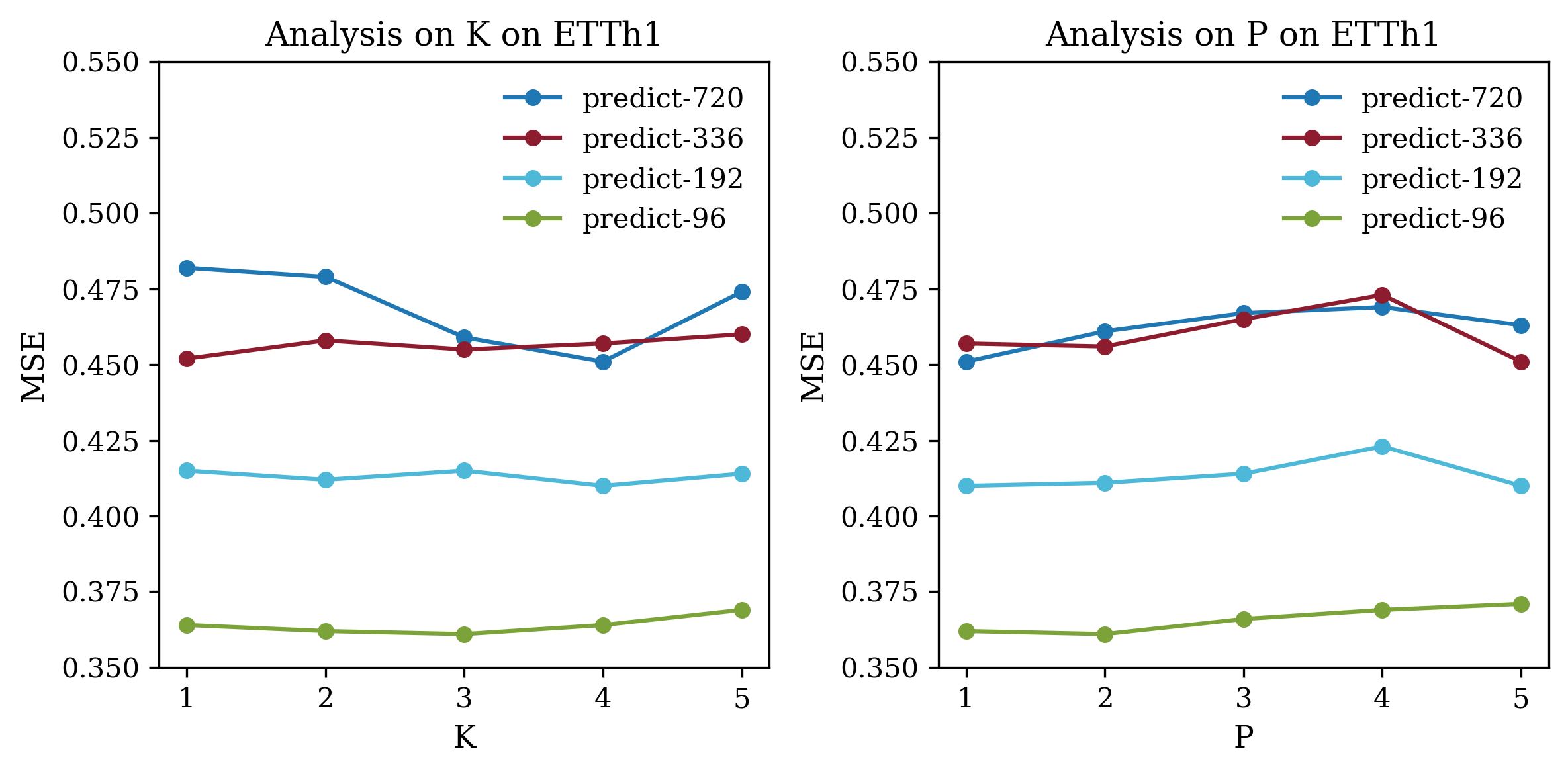}
  \caption{Analysis on the number of branches ($K$) and the number of components ($P$) on the ETTh1 dataset.}
  \label{fig:hyper_ETTh1_k_p}
\end{figure}

\textbf{Impact of hyperparameter $K$:}
We study the sensitivity of \NAME to the number of branches ($K$). 
As shown in the left panel of Figure \ref{fig:hyper_ETTh1_k_p}, the MSE on ETTh1 varies moderately as $K$ increases from 1 to 5, with no clear monotonic trend. 
Although a larger $K$ improves some horizons, e.g., $O=720$ at $K=4$, it does not consistently benefit all prediction lengths. 
For $O=336$, a small $K$ is already competitive, while the results for $O=96$ and $O=192$ remain relatively stable. 
This indicates that multiple branches provide useful capacity, but simply increasing $K$ does not guarantee better accuracy. 
Therefore, we adopt a moderate $K$ to balance accuracy and robustness.

\textbf{Impact of hyperparameter $P$:}
We study the sensitivity of \NAME to the number of components ($P$). 
As shown in the right panel of Figure \ref{fig:hyper_ETTh1_k_p}, the model shows moderate sensitivity to $P$ within the tested range. 
Overall, smaller $P$ values achieve better or comparable performance on most horizons. 
Increasing $P$ does not consistently reduce MSE for $O=96$, $O=192$, and $O=336$. 

These results suggest that a small number of components is usually sufficient, whereas larger $P$ may introduce unnecessary complexity. 
Thus, we adopt a relatively small $P$ to balance expressiveness and forecasting stability.

\textbf{Impact of hyperparamer $\psi$:}
To investigate the impact of the hyperparameter $\psi$, which defines the list of period lengths employed by the multi-branch architecture, we conduct experiments on three configurations with the same number of branches $K$ on the traffic dataset: (1) $\{24, 168\}$ capturing both daily and weekly patterns, (2) $\{24, 24\}$ using only daily periods, and (3) $\{168, 168\}$ using only weekly periods. Table~\ref{tabs:hyperparameter_period} reveals that the mixed-period configuration $\{24, 168\}$ consistently outperforms single-period variants across all prediction lengths, with the performance gap widening as the horizon extends. This validates that modeling on multiple periods simultaneously is crucial for long-term forecasting, as traffic patterns exhibit both intraday fluctuations and weekly periodicities. The robustness of $\{24, 168\}$ across varying horizons indicates that our multi-branch design effectively fuses complementary periodic information.

\renewcommand{\arraystretch}{1.0}
\begin{table}[htb]
\caption{Results of \NAME with varied $\psi$ on the Traffic dataset.}
\label{tabs:hyperparameter_period}

\resizebox{0.48\textwidth}{!}
{
\begin{tabular}{c|cc|cc|cc|cc}

\toprule

\multicolumn{1}{c|}{\multirow{1}{*}{Predict Length}} & \multicolumn{2}{c|}{\multirow{1}{*}{96}} & \multicolumn{2}{c|}{\multirow{1}{*}{192}} & \multicolumn{2}{c|}{\multirow{1}{*}{336}} & \multicolumn{2}{c}{\multirow{1}{*}{720}} \\


\cmidrule(lr){1-1} \cmidrule(lr){2-3} \cmidrule(lr){4-5} \cmidrule(lr){6-7} \cmidrule(lr){8-9}

$\psi$ & MSE & MAE & MSE & MAE & MSE & MAE & MSE & MAE \\

\toprule
$\{24,168\}$ & \textbf{0.457} & \textbf{0.283} & \textbf{0.463} & \textbf{0.282} & \textbf{0.478} & \textbf{0.288} & \textbf{0.514} & \textbf{0.310} \\

\midrule
$\{24,24\}$ & 0.460 & 0.284 & 0.464 & 0.284 & \textbf{0.478} & 0.289 & 0.517 & 0.312 \\

\midrule
$\{168,168\}$ & 0.460 & 0.285 & 0.465 & 0.284 & 0.479 & 0.289 & 0.517 & 0.313 \\

\bottomrule

\end{tabular}
}

\end{table}

\begin{figure}[t!]
  \centering
    \includegraphics[width=1\linewidth]{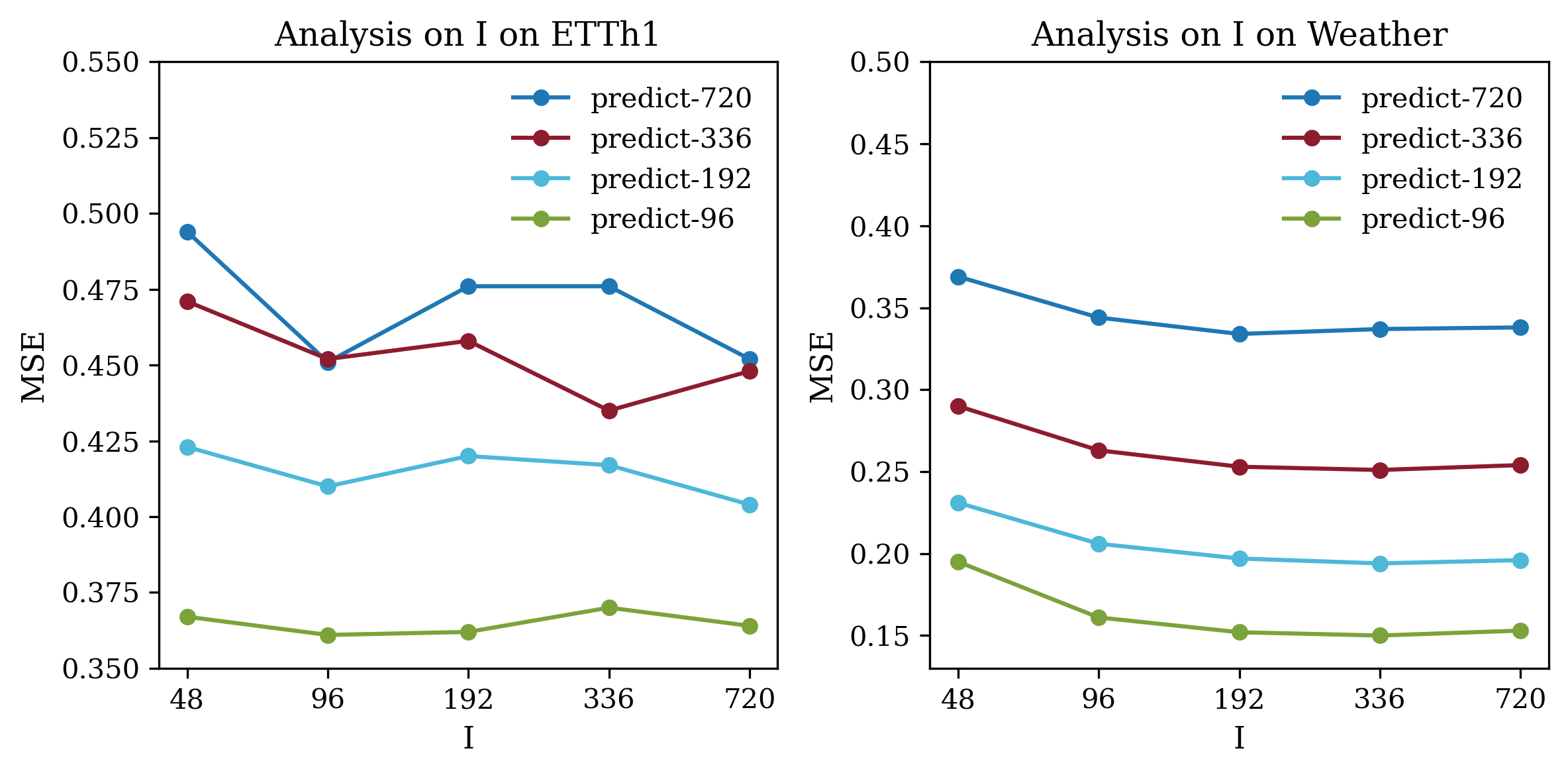}
  \caption{Analysis on the look back window ($I$) on the ETTh1 and the Weather dataset.}
  \label{fig:hyper_look_back}
\end{figure}

\textbf{Impact of look back window $I$:}
To investigate the impact of the input look-back window length $I$ on forecasting performance, we conduct experiments on the ETTh1 and the Weather dataset, as illustrated in Figure \ref{fig:hyper_look_back}. On the ETTh1 dataset, the performance exhibits a fluctuating but generally robust trend. On the Weather dataset, the MSE decreases as $I$ increases. 

\begin{figure}[t!]
  \centering
    \includegraphics[width=1\linewidth]{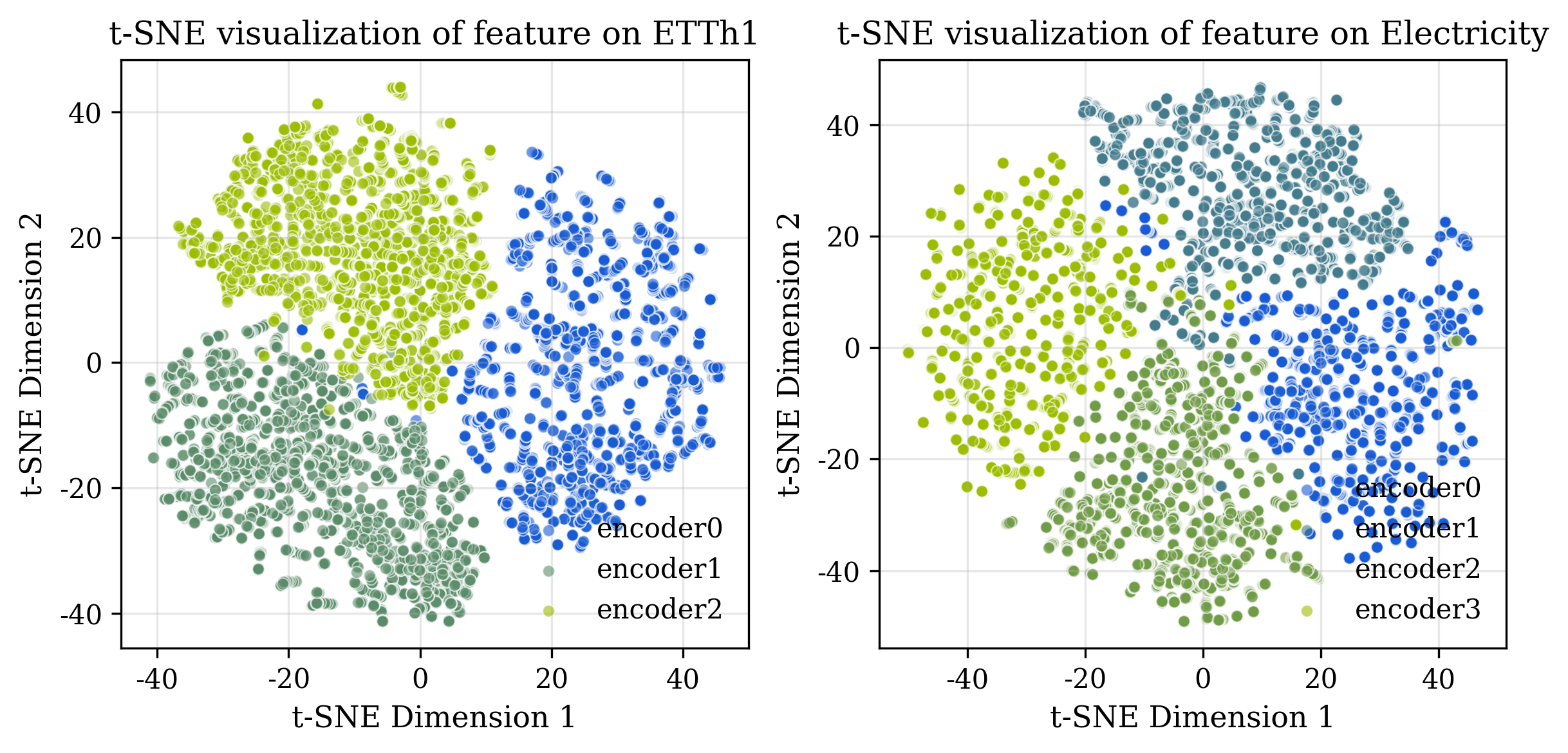}
  \caption{T-SNE visualization of the features from different encoders on the ETTh1 and Electricity dataset.}
  \label{fig:feature_tsne}
\end{figure}

\textbf{Feature Visualization:}
To intuitively understand the internal representations learned by our multi-branch encoders, we visualize the latent feature distributions extracted by different encoder branches using t-SNE~\cite{maaten2008visualizing}. We randomly sample feature vectors from the test sets of the ETTh1 and Electricity datasets and project them into a 2-dimensional space. As illustrated in Figure~\ref{fig:feature_tsne}, the features extracted by different encoders form highly distinct and well-separated clusters, validating that our multi-branch architecture does not simply learn redundant copies of the same information.

\section{Conclusions}

In this paper, we identified the fundamental limitations of topological mismatch and inefficient uniform representation inherent in existing 2D forecasting approaches.
To address this, we proposed \textbf{\NAME}, a novel framework that shifts the forecasting paradigm from point-wise regression to 2D generative rendering. 
By treating the future sequence as a latent surface to be reconstructed, \NAME explicitly enforces adjacency constraints through the inherent continuity of Gaussian kernels. 
To adapt 2D Gaussian Splatting to the temporal domain, we integrated the Multi-Basis Gaussian Kernel Generation to stabilize the optimization by synthesizing kernels from a fixed dictionary, alongside the Multi-Period Chronologically Continuous Rasterization to resolve topological mismatches between 2D grids and 1D sequences, thereby ensuring strict temporal continuity across periodic boundaries.
Extensive experiments on standard benchmark datasets demonstrate that \NAME achieves state-of-the-art or competitive performance. Our work not only validates the effectiveness of the rendering-based paradigm but also offers a fresh perspective on modeling complex coupled temporal dynamics.

\begin{acks}
This work is supported in part by the National Natural Science Foundation of China, under Grant (62302309, 62571298).
\end{acks}

\clearpage
\bibliographystyle{ACM-Reference-Format}
\bibliography{sample-base}

\clearpage
\appendix
\onecolumn


\section{Ablation Study}
\label{app:ablation}
In this section, we provide the full results of the ablation studies in Table \ref{tabs:ablation_all_full}. These results offer a detailed comparison of different model variants and help illustrate the individual contribution of each component to the overall performance.

\begin{table}[htbp]
\caption{Full results of ablation on different modules.}
\label{tabs:ablation_all_full}

\resizebox{\textwidth}{!}
{
\begin{tabular}{c|c|c|cc|cc|cc|cc|cc|cc}

\toprule

\multicolumn{1}{c|}{\multirow{2}{*}{Module}} &
\multicolumn{1}{c|}{\multirow{2}{*}{Method}} &
\multicolumn{1}{c|}{\multirow{2}{*}{Length}} &
\multicolumn{2}{c|}{ETTh1} &
\multicolumn{2}{c|}{ETTm1} &
\multicolumn{2}{c|}{ETTh2} &
\multicolumn{2}{c|}{ETTm2} &
\multicolumn{2}{c|}{Weather} &
\multicolumn{2}{c}{Electricity} \\

\cmidrule(lr){4-5}
\cmidrule(lr){6-7}
\cmidrule(lr){8-9}
\cmidrule(lr){10-11}
\cmidrule(lr){12-13}
\cmidrule(lr){14-15}

& & & MSE & MAE & MSE & MAE & MSE & MAE & MSE & MAE & MSE & MAE & MSE & MAE \\

\midrule

\multirow{4}{*}{Full Model}
& \multirow{4}{*}{Origin} & 96  & \textbf{0.361} & \textbf{0.386} & \textbf{0.307} & \textbf{0.347} & \textbf{0.281} & \textbf{0.329} & \textbf{0.169} & \textbf{0.251} & \textbf{0.161} & \textbf{0.199} & \textbf{0.151} & \textbf{0.237} \\
&                         & 192 & \textbf{0.410} & \textbf{0.419} & \textbf{0.354} & \textbf{0.373} & \textbf{0.359} & \textbf{0.380} & \textbf{0.239} & \textbf{0.293} & \textbf{0.206} & \textbf{0.242} & \textbf{0.164} & \textbf{0.250} \\
&                         & 336 & \textbf{0.452} & \textbf{0.440} & \textbf{0.383} & \textbf{0.397} & \textbf{0.401} & \textbf{0.416} & \textbf{0.298} & \textbf{0.332} & \textbf{0.263} & \textbf{0.286} & 0.184 & 0.270 \\
&                         & 720 & \textbf{0.451} & \textbf{0.451} & \textbf{0.443} & \textbf{0.433} & \textbf{0.410} & \textbf{0.431} & \textbf{0.396} & \textbf{0.391} & \textbf{0.344} & \textbf{0.337} & 0.224 & 0.304 \\

\midrule

\multirow{12}{*}{Feature Extraction}
& \multirow{4}{*}{Linear} & 96  & 0.392 & 0.401 & 0.323 & 0.357 & 0.289 & 0.336 & 0.179 & 0.261 & 0.169 & 0.212 & 0.168 & 0.253 \\
&                         & 192 & 0.442 & 0.427 & 0.366 & 0.379 & 0.379 & 0.395 & 0.247 & 0.303 & 0.217 & 0.253 & 0.176 & 0.260 \\
&                         & 336 & 0.486 & 0.447 & 0.401 & 0.403 & 0.411 & 0.420 & 0.304 & 0.339 & 0.272 & 0.292 & 0.191 & 0.276 \\
&                         & 720 & 0.497 & 0.479 & 0.470 & 0.441 & 0.419 & 0.437 & 0.402 & 0.396 & 0.349 & 0.342 & 0.232 & 0.309 \\

\cmidrule(lr){2-15}

& \multirow{4}{*}{MLP} & 96  & 0.377 & 0.392 & 0.323 & 0.357 & 0.288 & 0.333 & 0.175 & 0.256 & 0.165 & 0.205 & 0.157 & 0.243 \\
&                     & 192 & 0.430 & 0.421 & 0.365 & 0.379 & 0.366 & 0.383 & 0.241 & 0.299 & 0.213 & 0.248 & 0.168 & 0.254 \\
&                     & 336 & 0.471 & 0.441 & 0.400 & 0.402 & 0.407 & 0.419 & 0.299 & 0.337 & 0.268 & 0.289 & 0.189 & 0.276 \\
&                     & 720 & 0.485 & 0.467 & 0.471 & 0.444 & 0.414 & 0.433 & 0.397 & 0.395 & 0.348 & 0.341 & 0.230 & 0.309 \\

\cmidrule(lr){2-15}

& \multirow{4}{*}{Plain CNN} & 96  & 0.371 & 0.387 & 0.326 & 0.361 & \textbf{0.281} & 0.331 & 0.174 & 0.255 & 0.167 & 0.208 & 0.163 & 0.249 \\
&                           & 192 & 0.440 & 0.425 & 0.369 & 0.380 & 0.363 & 0.383 & 0.241 & 0.298 & 0.217 & 0.251 & 0.172 & 0.256 \\
&                           & 336 & 0.479 & 0.446 & 0.399 & 0.400 & 0.410 & 0.419 & 0.301 & 0.338 & 0.271 & 0.291 & 0.192 & 0.278 \\
&                           & 720 & 0.479 & 0.465 & 0.467 & 0.438 & 0.414 & 0.432 & 0.397 & 0.395 & 0.350 & 0.344 & 0.233 & 0.312 \\

\midrule

\multirow{4}{*}{Kernel Generation}
& \multirow{4}{*}{w/o Multi-Basis} & 96  & 0.596 & 0.492 & 0.627 & 0.514 & 0.489 & 0.388 & 0.183 & 0.262 & 0.187 & 0.219 & 0.852 & 0.764 \\
&                                  & 192 & 0.494 & 0.450 & 0.595 & 0.546 & 0.525 & 0.451 & 0.565 & 0.333 & 0.264 & 0.283 & 0.863 & 0.764 \\
&                                  & 336 & 0.583 & 0.556 & 0.729 & 0.566 & 0.469 & 0.452 & 0.340 & 0.365 & 0.301 & 0.316 & 0.891 & 0.765 \\
&                                  & 720 & 0.534 & 0.501 & 0.827 & 0.580 & 0.553 & 0.463 & 0.462 & 0.420 & 0.373 & 0.369 & 0.901 & 0.774 \\

\midrule

\multirow{8}{*}{Rendering}
& \multirow{4}{*}{Linear} & 96  & 0.379 & 0.394 & 0.315 & 0.358 & 0.286 & 0.332 & 0.175 & 0.256 & 0.163 & 0.203 & 0.155 & 0.241 \\
&                         & 192 & 0.416 & 0.423 & 0.363 & 0.382 & 0.363 & 0.384 & 0.241 & 0.299 & 0.210 & 0.245 & 0.167 & 0.253 \\
&                         & 336 & 0.457 & 0.443 & 0.392 & 0.405 & 0.422 & 0.424 & 0.303 & 0.338 & 0.265 & \textbf{0.286} & 0.183 & 0.270 \\
&                         & 720 & 0.509 & 0.479 & 0.447 & 0.438 & 0.419 & 0.437 & 0.399 & 0.395 & 0.345 & 0.338 & 0.223 & 0.303 \\

\cmidrule(lr){2-15}

& \multirow{4}{*}{MLP} & 96  & 0.363 & 0.389 & 0.311 & 0.350 & 0.291 & 0.336 & 0.172 & 0.254 & 0.162 & 0.202 & 0.153 & 0.241 \\
&                     & 192 & 0.413 & 0.423 & 0.355 & 0.376 & 0.362 & 0.381 & 0.241 & 0.299 & 0.207 & 0.243 & 0.166 & 0.252 \\
&                     & 336 & 0.465 & 0.443 & 0.385 & 0.401 & 0.423 & 0.424 & 0.300 & 0.336 & 0.264 & 0.287 & \textbf{0.181} & \textbf{0.268} \\
&                     & 720 & 0.453 & 0.453 & \textbf{0.443} & 0.435 & 0.414 & 0.432 & 0.398 & 0.393 & 0.345 & 0.339 & \textbf{0.216} & \textbf{0.299} \\

\midrule

\multirow{8}{*}{Aggregation}
& \multirow{4}{*}{Average} & 96  & 0.366 & 0.390 & 0.311 & 0.350 & 0.288 & 0.335 & 0.177 & 0.257 & 0.171 & 0.207 & 0.163 & 0.245 \\
&                          & 192 & 0.421 & 0.426 & 0.364 & 0.380 & 0.363 & 0.382 & 0.242 & 0.300 & 0.215 & 0.248 & 0.173 & 0.255 \\
&                          & 336 & 0.468 & 0.446 & 0.391 & 0.401 & 0.410 & 0.419 & 0.303 & 0.339 & 0.272 & 0.291 & 0.195 & 0.277 \\
&                          & 720 & 0.457 & 0.457 & 0.447 & 0.434 & 0.418 & 0.435 & 0.397 & 0.394 & 0.351 & 0.340 & 0.232 & 0.310 \\

\cmidrule(lr){2-15}

& \multirow{4}{*}{Channel Agnostic} & 96  & \textbf{0.361} & \textbf{0.386} & 0.310 & 0.349 & 0.287 & 0.333 & 0.174 & 0.255 & 0.171 & 0.208 & 0.163 & 0.244 \\
&                                  & 192 & 0.411 & \textbf{0.419} & 0.357 & 0.374 & 0.364 & 0.382 & 0.241 & 0.298 & 0.217 & 0.248 & 0.172 & 0.254 \\
&                                  & 336 & \textbf{0.452} & \textbf{0.440} & 0.387 & 0.398 & 0.416 & 0.420 & 0.300 & 0.337 & 0.273 & 0.289 & 0.193 & 0.276 \\
&                                  & 720 & \textbf{0.451} & \textbf{0.451} & 0.449 & 0.436 & 0.417 & 0.434 & \textbf{0.396} & 0.393 & 0.349 & 0.339 & 0.232 & 0.309 \\

\bottomrule

\end{tabular}
}

\end{table}





\section{Loss Function}
To train \NAME, we employ a hybrid loss function designed to optimize the intensity and basis mixing weights of the Gaussian kernels ~\cite{huber1992robust}. Given the rendered prediction $\hat{\mathbf{Y}}$ and the ground truth $\mathbf{Y}$, our objective minimizes a weighted combination of Mean Squared Error (MSE) and Mean Absolute Error (MAE):
\begin{align}
    \mathcal{L} = \lambda \cdot \mathcal{L}_{MSE}(\mathbf{Y}, \hat{\mathbf{Y}}) + (1 - \lambda) \cdot \mathcal{L}_{MAE}(\mathbf{Y}, \hat{\mathbf{Y}}),
\end{align}
where $\lambda \in [0, 1]$ is a hyperparameter balancing the two terms.

\section{Implementation Details}
All experiments are implemented in PyTorch~\cite{paszke2019pytorch} and run on NVIDIA GeForce RTX 3090 24GB GPUs.
We adopt the Adam optimizer~\cite{adam} with an initial learning rate in $\{10^{-3}, 10^{-4}\}$ and a batch size from $\{16, 32\}$. 
Crucially, for our \NAME, the hyperparameter $\psi$ (period length for 2D Gaussian rasterization) is selected based on the domain knowledge of each dataset. For hourly datasets like ETTh and Electricity, $\psi_k=24$ is used to capture daily cycles. The number of branches $K$ and the number of components in each branch $P$ are tuned via grid search.
Detailed model configuration information is presented in Table~\ref{tabs:configuration}.
\begin{table}[htbp]
\caption{Experiment configuration of \NAME.}
\label{tabs:configuration}

\resizebox{0.75\textwidth}{!}
{
\begin{tabular}{c|c|c|c|c|c|c|c}

\toprule

Configurations & ETTh1 & ETTh2 & ETTm1 & ETTm2 & Weather & Electricity & Traffic \\

\toprule

$\text{Set}(\psi)$ & $\{24\}$ & $\{24\}$ & $\{96\}$ & $\{96\}$ & $\{144\}$ & $\{24\}$ & $\{24,168\}$ \\

$\lambda$ & 0.5 & 0.0 & 0.5 & 0.5 & 0.5 & 0.5 & 0.5 \\

Learning Rate & $10^{-4} $ & $10^{-4} $ & $10^{-3} $ & $10^{-4} $ & $10^{-4} $ & $10^{-3} $ & $10^{-4}$ \\

Batch Size & 32 & 16 & 16 & 32 & 16 & 16 & 16 \\

\bottomrule

\end{tabular}
}

\end{table}


\section{Showcases}
\label{app:visualization}
\begin{figure*}[t!]
      \centering
      \includegraphics[width=\linewidth]{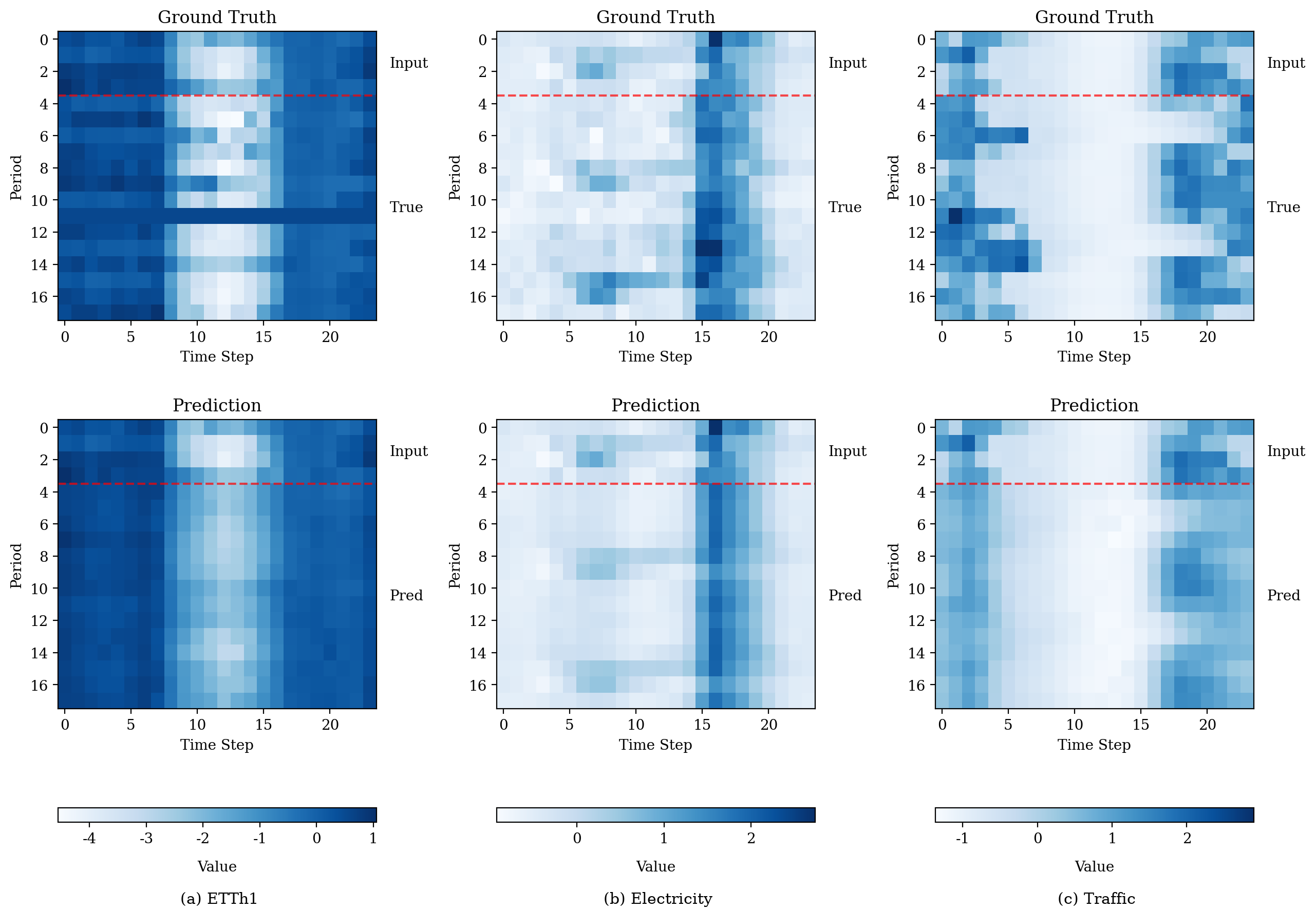}
      \caption{Showcases for 2d variations from ETTh1, Electricity and Traffic by \NAME under the input-96-predict-336 settings.}
        \label{fig:2dvisualization}
    \end{figure*}

We assess model performance by visualizing the final prediction on the test sets across different datasets, which provides an intuitive comparison between predictions and ground truth. We also visualize the 2D variations of the real and predicted data, as shown in Figure~\ref{fig:2dvisualization}.

    \begin{figure*}[t!]
      \centering
      \includegraphics[width=0.95\linewidth]{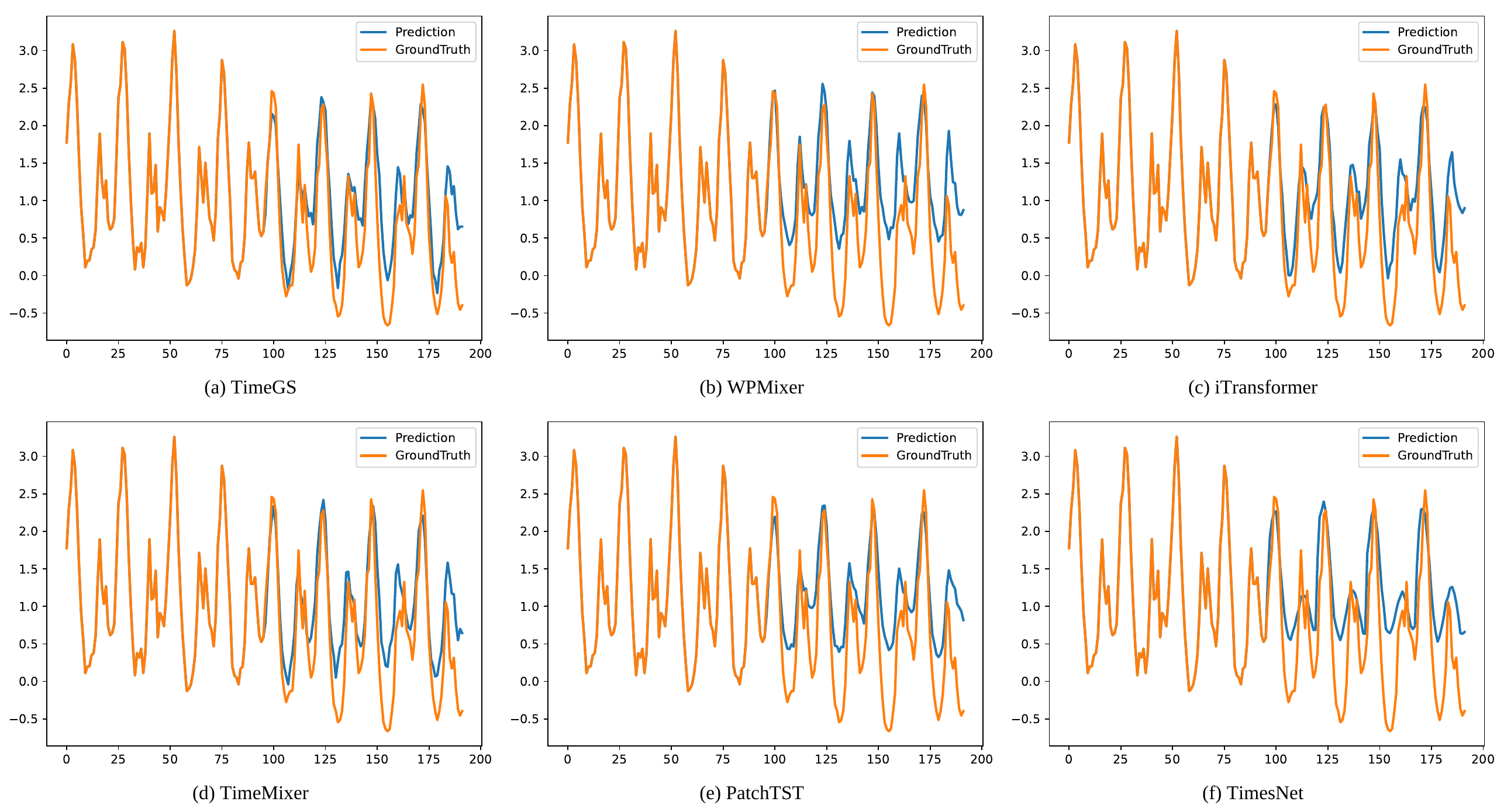}
      \caption{Showcases from ETTh1 by different models under the input-96-predict-96 settings.}
        \label{fig:visualization_ETTh1}
    \end{figure*}
    
    \begin{figure*}[t!]
      \centering
      \includegraphics[width=0.95\linewidth]{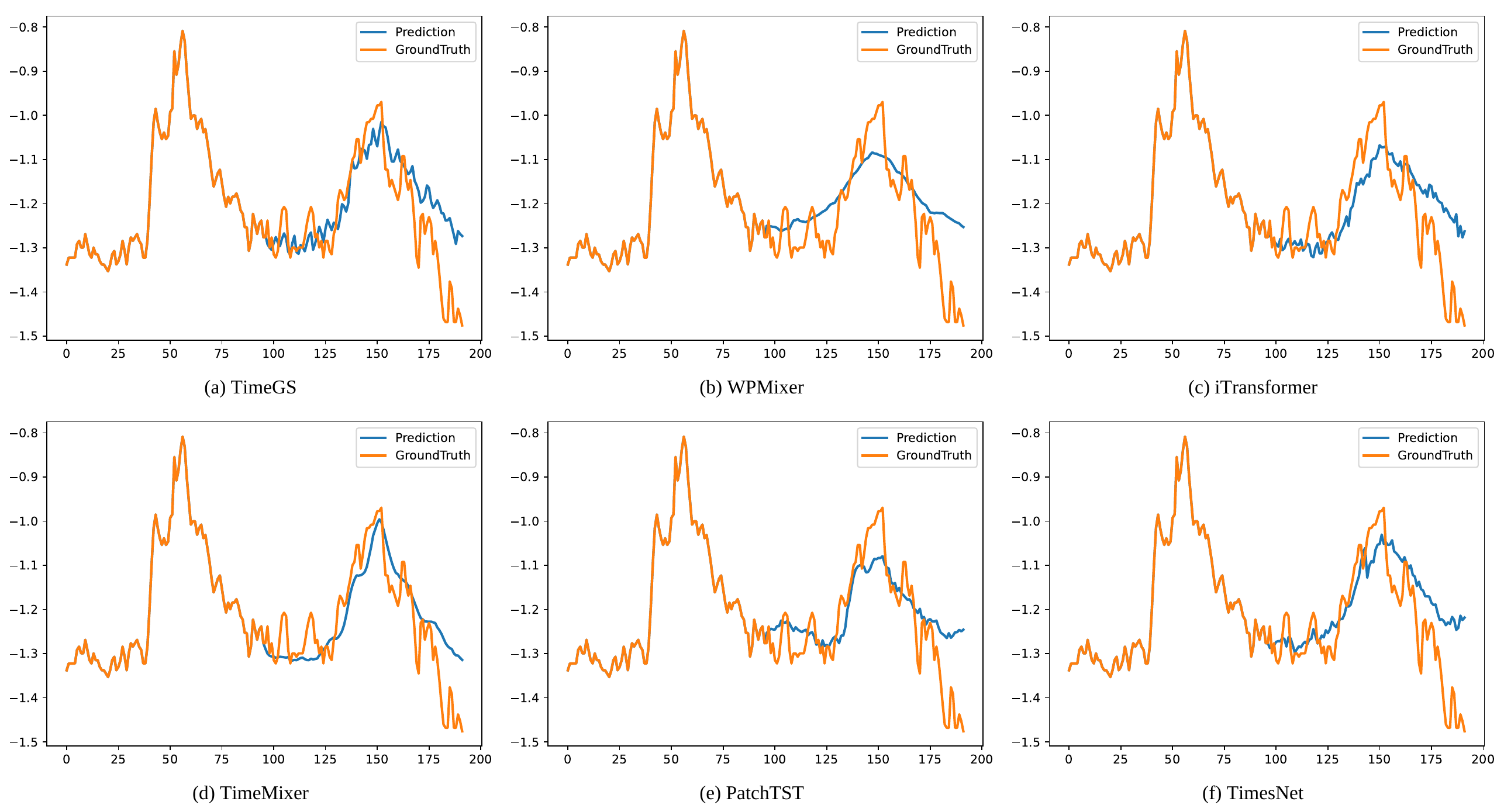}
      \caption{Showcases from ETTm1 by different models under the input-96-predict-96 settings.}
        \label{fig:visualization_ETTm1}
    \end{figure*}
    
    \begin{figure*}[t!]
      \centering
      \includegraphics[width=0.95\linewidth]{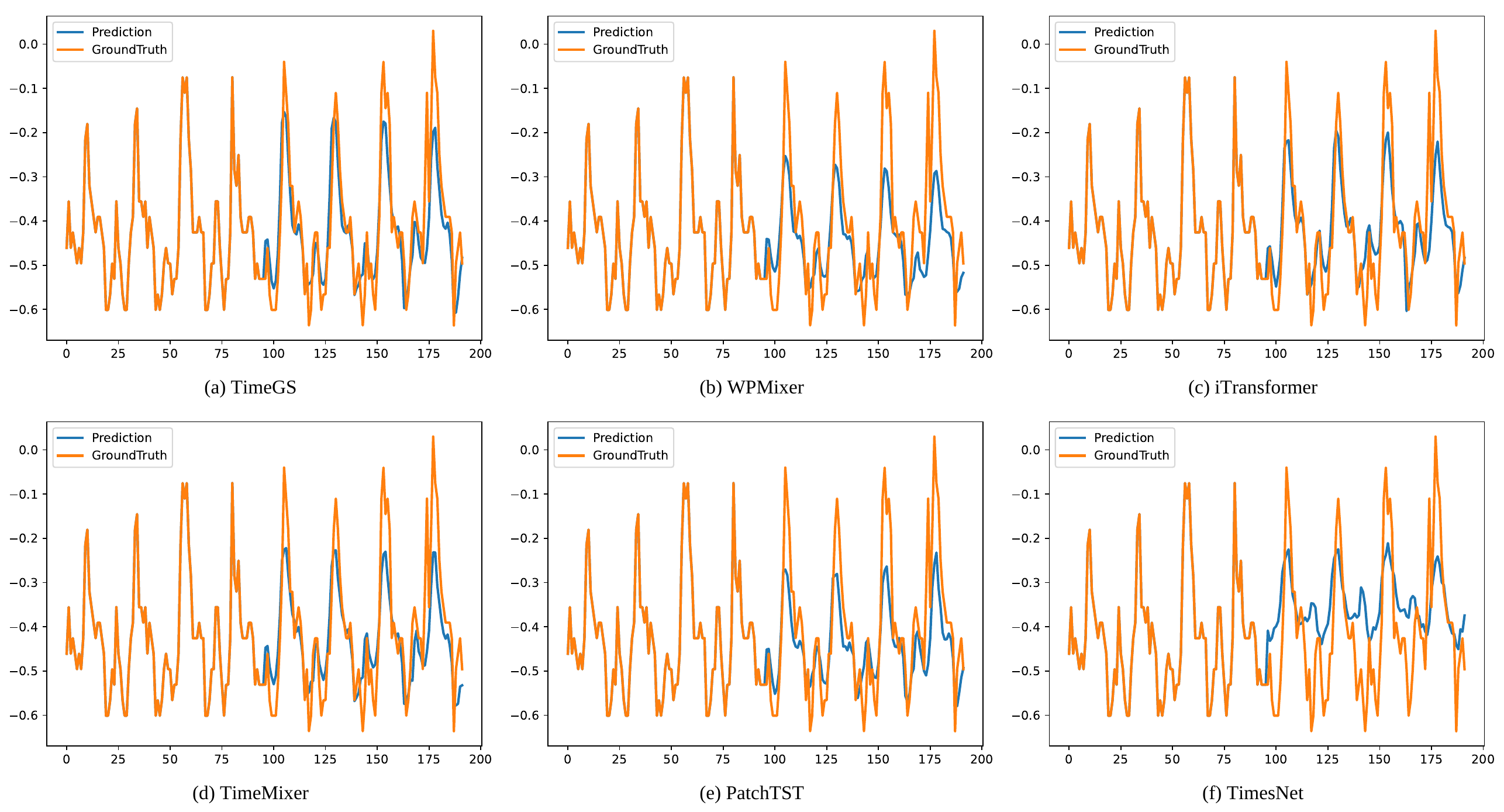}
      \caption{Showcases from Electricity by different models under the input-96-predict-96 settings.}
        \label{fig:visualization_ECL}
    \end{figure*}
    
    \begin{figure*}[t!]
      \centering
      \includegraphics[width=0.95\linewidth]{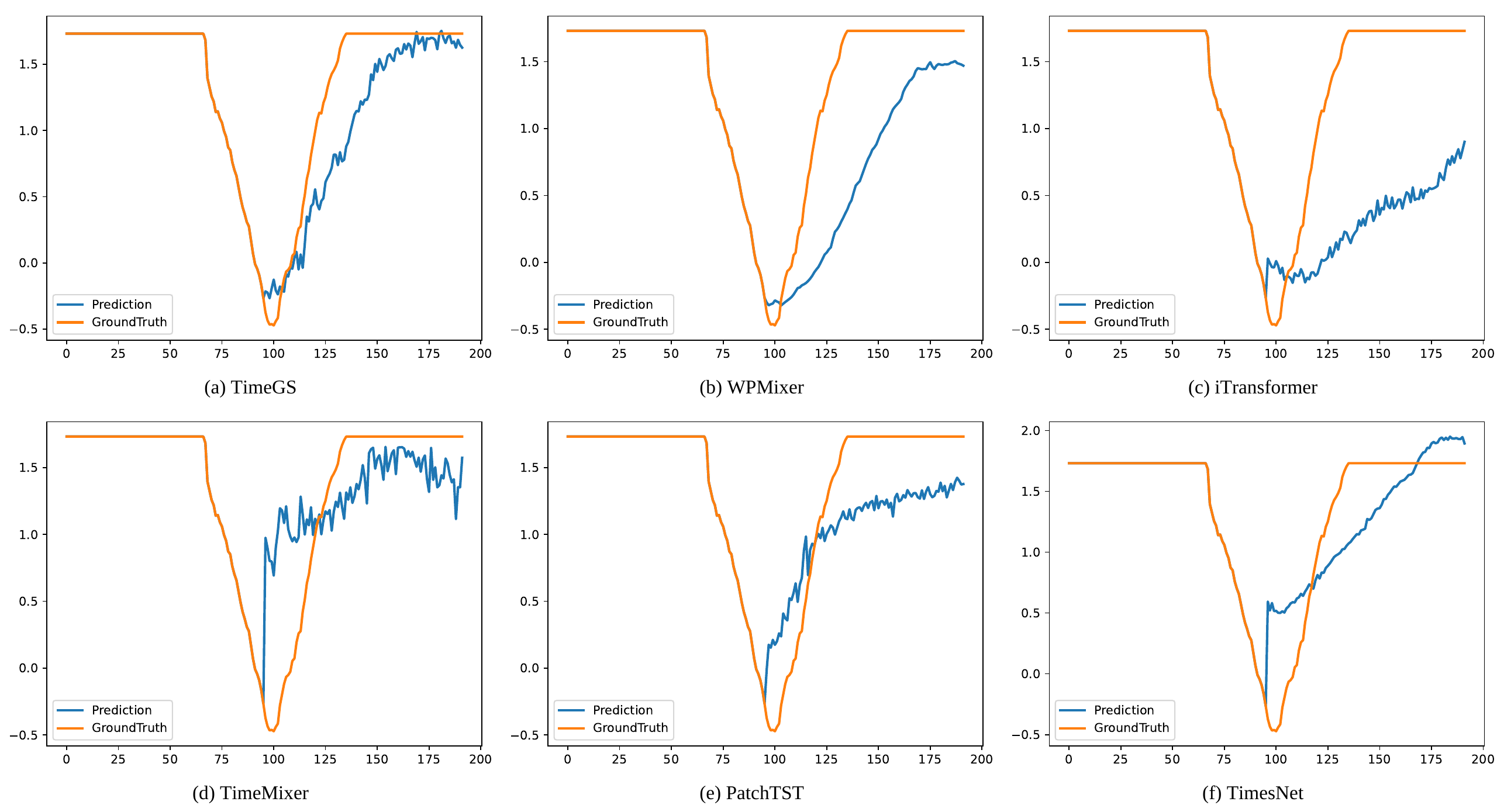}
      \caption{Showcases from Weather by different models under the input-96-predict-96 settings.}
        \label{fig:visualization_weather}
    \end{figure*}

\end{document}